\documentclass[sigconf,screen,authorversion]{acmart}

\AtBeginDocument{%
  \providecommand\BibTeX{{%
    \normalfont B\kern-0.5em{\scshape i\kern-0.25em b}\kern-0.8em\TeX}}}

\copyrightyear{2023}
\acmYear{2023}
\setcopyright{rightsretained}
\acmConference[SC-W 2023]{Workshops of The International Conference on High Performance Computing, Network, Storage, and Analysis}{November 12--17, 2023}{Denver, CO, USA}
\acmBooktitle{Workshops of The International Conference on High Performance Computing, Network, Storage, and Analysis (SC-W 2023), November 12--17, 2023, Denver, CO, USA}
\acmDOI{10.1145/3624062.3626078}
\acmISBN{979-8-4007-0785-8/23/11}

\usepackage[labelfont=sc,font=bf,skip=10pt]{caption}
\usepackage[labelfont=up,skip=1pt]{subcaption}
\usepackage{booktabs}
\usepackage{multirow}
\usepackage{cases}
\usepackage{empheq}
\usepackage{xfrac}
\usepackage{cleveref}

\graphicspath{{figure/}{figures/}}
\DeclareMathOperator*{\argmin}{argmin}
\DeclareMathOperator{\sech}{sech}
\newcommand{\tabhead}[1]{{\bfseries#1}}
\newcommand{\md}{\text{d}}
\DeclareMathOperator{\e}{e}

\begin{document}

%% Title
\title[Mesh-free differentiable programming for optimal control]{A comparison of mesh-free differentiable programming and data-driven strategies for optimal control under PDE constraints}

% \titlenote{Produces the permission block, and copyright information}
% \subtitle{Paper \#, XXX pages}

%% Authors
\author{Roussel D. Nzoyem Ngueguin}
\orcid{0009-0008-2019-9973}
\email{rd.nzoyemngueguin@bristol.ac.uk}
\affiliation{
  \department{School of Computer Science}
  \institution{University of Bristol}
  \city{Bristol}
  \country{UK}
  \postcode{BS8 1UB}
}

\author{David A.W.\ Barton}
\orcid{0000-0002-0595-4239}
\email{David.Barton@bristol.ac.uk}
\affiliation{
  \department{School of Engineering Mathematics and Technology}
  \institution{University of Bristol}
  \city{Bristol}
  \country{UK}
  \postcode{BS8 1TW}
}

\author{Tom Deakin}
\orcid{0000-0002-6439-4171}
\email{tom.deakin@bristol.ac.uk}
\affiliation{
  \department{School of Computer Science}
  \institution{University of Bristol}
  \city{Bristol}
  \country{UK}
  \postcode{BS8 1UB}
}

%% Short authors in page headers

%% Abstract
\begin{abstract}

The field of Optimal Control under Partial Differential Equations (PDE) constraints is rapidly changing under the influence of Deep Learning and the accompanying automatic differentiation libraries. Novel techniques like Physics-Informed Neural Networks (PINNs) and Differentiable Programming (DP) are to be contrasted with established numerical schemes like Direct-Adjoint Looping (DAL). We present a comprehensive comparison of DAL, PINN, and DP using a general-purpose mesh-free differentiable PDE solver based on Radial Basis Functions. Under Laplace and Navier-Stokes equations, we found DP to be extremely effective as it produces the most accurate gradients; thriving even when DAL fails and PINNs struggle. Additionally, we provide a detailed benchmark highlighting the limited conditions under which any of those methods can be efficiently used. Our work provides a guide to Optimal Control practitioners and connects them further to the Deep Learning community.

\end{abstract}

%% Computing Classification System from http://dl.acm.org/ccs.cfm



%% Keywords

%% A "teaser" image
\begin{teaserfigure}
    \centering
  \includegraphics[width=.94\textwidth]{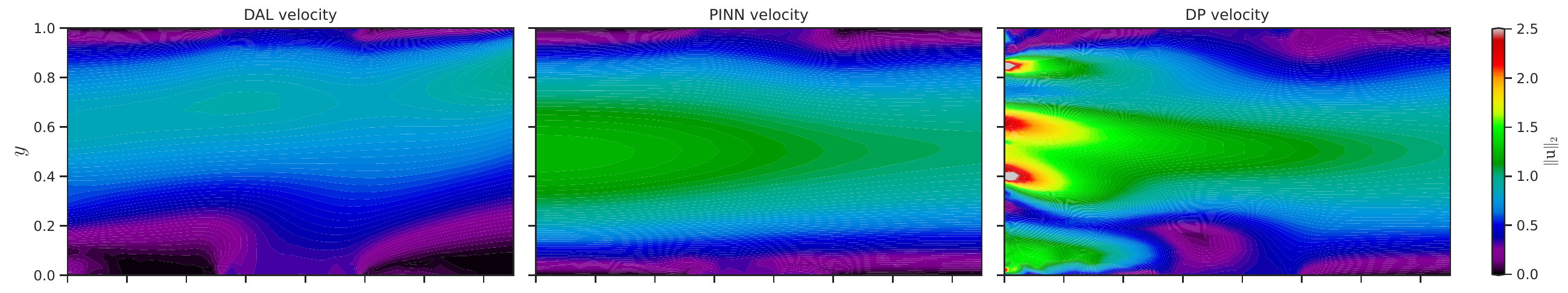}
  \includegraphics[width=.94\textwidth]{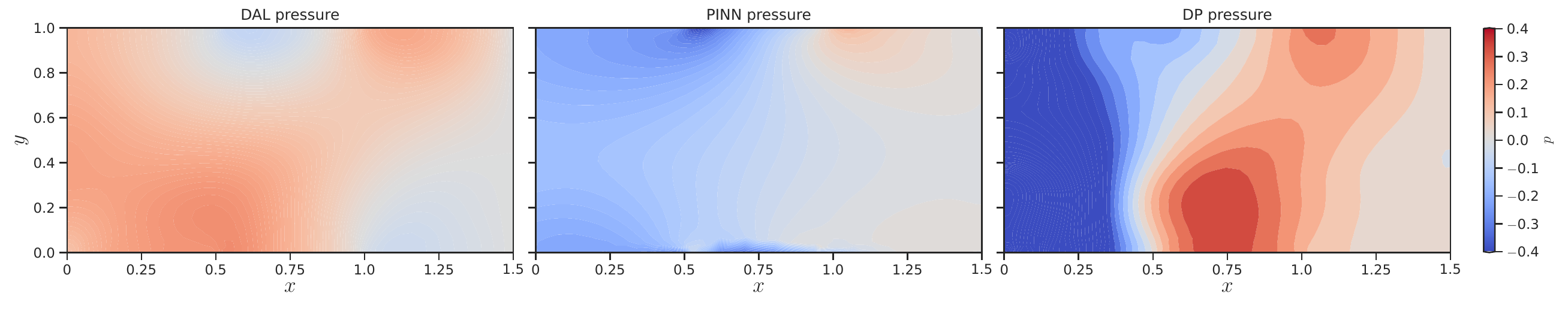}
  \caption{Qualitative evaluation of our  differentiable programming (DP) framework against direct-adjoint looping (DAL) and physics-informed neural networks (PINNs) when the goal is to control the inflow velocity at $x=0$ to achieve a parabolic outflow profile at $x=1.5$ given a cross-flow at the mid-point.}
  \Description{DP is better. Full stop.}
  \label{fig:teaser}
\end{teaserfigure}


%% Processes and builds the above
\maketitle

%% Main sections
\section{Introduction}
\label{sec.intro}

In the physical sciences, ordinary and partial differential equations (ODEs and PDEs) are widely used to model natural phenomena. Controlling the behaviour of systems governed by such equations is vital to many engineering disciplines. Optimal control (OC) is the task of minimising a cost objective $\mathcal{J}$ subject to specific dynamical constraints \cite{trelat2005controle}. In \cref{fig:teaser} for instance, given an input velocity $\mathbf{u}_{\text{in}}$, the flow in the channel is fully determined by the Navier-Stokes equations. OC formalises related optimisation processes and describes how actions on $\mathbf{u}_{\text{in}}$ affect $\mathbf{u}_{\text{out}}$ at the outlet. It helps find the input condition for a desired output. Nowadays, automated systems based on OC are part of everyday life: spacecraft fuel control \cite{bonnard2006mecanique}, population dynamics \cite{allegretto2011coexistence}, computer files transfer \cite{trelat2005controle}, social distancing and COVID-19 \cite{tsay2020modeling}, the list goes on. Like many optimisation problems, gradients are at the heart of OC.

% Lagrange's famous treatise that laid the foundation for Lagrange multipliers and adjoint-based sensitivity analysis dates back to 1853 \cite{lagrange1853mecanique}. Since then, numerous methods have been devised to evaluate the gradient of $\mathcal{J}$ with respect to a control variable $c$. One of the most established numerical algorithms is direct-adjoint looping (DAL) which has been successfully applied to problems in shape optimisation \cite{allaire2015review}, turbulent fluid flow \cite{nabi2019nonlinear}, and many more. Like most adjoint-based schemes, DAL works by solving, in addition to the original governing equations for the state $\mathbf{u}$, a second PDE for $\mathbf{\lambda}$ termed the \emph{adjoint state} problem. $\mathbf{u}$ and $\mathbf{\lambda}$ are then used iteratively to update $c$ until a stopping condition is met.

One of the most established numerical algorithms for OC is direct-adjoint looping (DAL) which has been successfully applied to problems in shape optimisation \cite{allaire2015review}, turbulent fluid flow \cite{nabi2019nonlinear}, and many more. It builds on Lagrange's famous 1853 treatise~\cite{lagrange1853mecanique}, which laid the foundation for Lagrange multipliers and adjoint-based sensitivity analysis. At its core, it iteratively evaluates the gradient of $\mathcal{J}$ with respect to a control variable $c$, until a stopping condition is met. Like most adjoint-based schemes, DAL works by solving a second PDE for $\mathbf{\lambda}$ termed the \emph{adjoint} problem, in addition to the original (direct) governing equations.

% \paragraph{The problem with DAL} 
DAL is a powerful OC approach, but it has significant drawbacks. Just setting up the adjoint problem in DAL is a non-trivial task that often overshadows the computational effort of interest. For simple OC problems with complex PDEs, it is hard to justify the use of DAL and its significant derivation overhead. For extremely hard problems like engineering design applications, we often have a set of constraints which must be satisfied in addition to the governing equations. Some of these may be geometric and others may be overly entangled with the PDE variables. The \emph{soft constraints} approach is typically employed to deal with those, thus eroding DAL's main selling points \cite{giles2000introduction}.

Physics-Informed Neural Networks (PINNs)\footnote{In this mesh-free context, our work simply explores the usability of PINNs for OC, and is therefore limited to their original (vanilla) form.} \cite{raissi2019physics,cuomo2022scientific} are among a new wave of methods for solving forward and inverse problems that followed the renaissance of Deep Neural Networks. The breathtaking success of Deep Learning is fuelled by the availability of large volumes of data and tensor-oriented accelerators, beautifully brought together with the \emph{backpropagation} algorithm \cite{rumelhart1986learning,werbos1990backpropagation}. This algorithm for exact and efficient computation of gradients underpins transformative discoveries in fields such as computer vision \cite{hinton2012imagenet}, natural language processing \cite{vaswani2017attention}, and scientific machine learning \cite{cuomo2022scientific} to state just those. The intuition behind backpropagation goes back to Lagrange multipliers and its adjoint states \cite{lagrange1853mecanique}, repackaged more generally as reverse-mode automatic differentiation (AD) \cite{griewank2008evaluating}. Recognising its wide applicability outside Deep Learning, several works have leveraged AD to propagate gradients through an entire PDE's discretised solver: $\phi-$Flow \cite{holl2020learning} and Mitsuba \cite{nimier2019mitsuba} geared towards computer graphics, \texttt{JAX-CFD} \cite{bezgin2023jax} for fluid dynamics, \texttt{JAX-MD} \cite{schoenholz2019jax} for molecular dynamics, and many more specialised differentiable solvers for geosciences \cite{shen2023differentiable}, robotics \cite{howell2022dojo}, etc.

% \paragraph{The problem with PINNs} 
Like DAL, PINNs in their original form \cite{raissi2019physics} are subject to crippling weaknesses. Surrogate models built with them struggle to generalise to unseen scenarios, and they become harder to train in multi-objective settings. It is particularly hard for PINNs to approximate discontinuous PDE solutions which can develop from smooth initial conditions over time. In Computational Fluid Dynamics, for instance, PINNs routinely fail to learn multiscale, chaotic, or turbulent behaviours, hence limiting their appeal for robust engineering applications \cite{cuomo2022scientific}. The main reason behind these issues is the \emph{variational crime} \cite{Beguinet2023}, which states that by minimising the residual's norm in their loss functions, PINNs demand more regularity than what the theory would typically allow.

In light of this plethora of numerical methods in modern-day control settings, it is important to know the strengths and weaknesses of each method, along with their new use cases outside their original domains. OC practitioners need a robust yet flexible tool to quickly prototype models and control them under various conditions. Furthermore, educating oneself on the mechanisms and future trends surrounding these methods should be beneficial to both OC and Deep Learning communities.

One potential solution to the problems of DAL and PINNs is differentiable programming (DP)\footnote{Another popular perspective is to view DP as a broad generalisation encompassing DAL on one end of the spectrum and PINNs on the other end.}, specifically its \emph{discretise-then-optimise} (DTO) approach to leveraging AD. In the Python ecosystem, several libraries can enable DP. JAX \cite{jax2018github} is one such library that exposes low-level composable transformations like AD, vectorisation, parallelisation, and just-in-time compilation to XLA kernels \cite{sabne2020xla} for various hardware accelerators. In this work, we use JAX not only to implement PINNs, but also to build \texttt{$\mathbb{U}$pdec}: an end-to-end differentiable PDE solver suitable for optimal control \cite{nzoyem2023updec}. The resulting framework for DP is compared to DAL and PINN for accuracy and computational performance on two problems in engineering settings: the Laplace problem on a unit square, and the fluid flow in a channel governed by the Navier-Stokes equations.

Before considering control, the underlying PDEs must be simulated with either of the following two categories of techniques: \emph{with} or \emph{without} a mesh. Using a mesh confers an important inductive bias, namely for discretising spatial differential operators. In contrast, mesh-free\footnote{In CFD, Smoothed Particle Hydrodynamics (SPH) is another mesh-free method particularly attractive for its Lagrangian nature.} methods do not require such structure, and are therefore attractive when the geometry is complex. Radial Basis Functions (RBFs) are interpolants that allow for robust, intuitive, and mesh-free simulation of a large variety of PDEs \cite{kansa1990multiquadrics}. RBFs are the backbone of our DAL and DP numerical schemes. The choice of such an inherently mesh-free method is further motivated by a desire to test our RBF-based implementations against the data-driven but equally mesh-free PINN.

Our contribution is twofold. First, our comparative study of DAL, DP, and PINN in a unified mesh-free context is, to our knowledge, the first of its kind. Our second contribution lies in identifying RBFs as an effective method not only for PDE simulation, but also for enhanced optimisation via the user-friendly \texttt{$\mathbb{U}$pdec} differentiable programming framework \cite{nzoyem2023updec}.

% Our main contribution lies in identifying RBFs as a viable simulation method not only for PDE simulation, but also for enhanced optimisation via differentiable programming. To the best of the authors' knowledge, this is the only comprehensive study comparing both established methods like DAL with up-and coming alternatives based on AD.

\section{Methods}
\label{sec.method}

\subsection{Radial Basis Functions}
\label{subsec:rbf}

Radial Basis Functions have a long history rooted in approximation theory \cite{cheney1966introduction,powell1981approximation}. They were revived in the last 30 years, particularly for their use as activation layers in Neural Networks \cite{orr1996introduction}. Building on their \emph{universal} and \emph{best} approximation theorems \cite{acosta1995radial,park1991universal}, the expressive power of RBFs is typically used to model PDEs of the form 
\begin{subequations}    \label{eq:rbfpde}
\begin{empheq} [left=\empheqlbrace] {align}
    \mathcal{D}(u) &= q  &&\text{in } \Omega \label{eq:pde1} \\
    u &= q_d  &&\text{on } \Gamma_d \label{eq:pde2}\\
    \frac{\partial u}{\partial \mathbf{n}} &= q_n  &&\text{on } \Gamma_n \label{eq:pde3}\\
    \frac{\partial u}{\partial \mathbf{n}} + \beta u &= q_r,  &&\text{on } \Gamma_r \label{eq:pde4}
\end{empheq}
\end{subequations}
where $\Omega \in \mathbb{R}^d $ with boundary $\partial \Omega = \Gamma_d \cup \Gamma_n \cup \Gamma_r$.\footnote{The subscripts $d$, $n$ and $r$ stand for the three major types of Boundary conditions: Dirichlet, Neumann, and Robin, respectively.} $q$, $q_d$, $q_n$, $q_r$, and $\beta$ are known fields defined on $\Omega$, $\Gamma_d$, $\Gamma_n$, $\Gamma_r$, and $\Gamma_r$ respectively. The vector $\mathbf{n}$ is the outward-facing normal along the corresponding boundary. $\mathcal{D}$ is a (combination of) linear differential operator(s) applied to the unknown scalar field $u$ interpolated as
\begin{align} \label{eq:rbffield}
    \hat u(\mathbf{x}) = \sum_{j=1}^{N} \lambda_j \phi( \Vert \mathbf{x} - \mathbf{x}_j \Vert_2 ) + \sum_{j=1}^{M} \gamma_j P_j(\mathbf{x}),
\end{align} 
where $\phi$ is a radial basis function (RBF): a real-valued function of the Euclidean distance $ \Vert \cdot \Vert_2$. The nodal points $\{\mathbf{x}_j\}_{j=1,\ldots, N}$ represent centres (or centroids) of the RBFs, arbitrarily scattered in $\Omega$ and on $\partial \Omega$. $\{P_j\}_{j=1,\ldots,M}$ are appended polynomials as suggested in the RBF-FD framework \cite{tolstykh2000using}. Solving \cref{eq:rbfpde} means collocating $\hat u$ at nodes $\{\mathbf{x}_i\}_{i=1,\ldots, N_{points}}$ (which can be the same as the centroids), then applying $\mathcal{D}$, resulting in a linear system of simultaneous equations whose solution yields the coefficients $\lambda_j$ and $\gamma_j$.

Under certain conditions, RBFs are known to suffer from the Runge phenomenon \cite{fornberg2007runge,fornberg2008stable}, leading to poor approximation near $\partial \Omega$. The question of boundary handling with RBF has always been tricky, with some of the most impactful applications of RBF bypassing it by using background meshes \cite{shahane2020ahighorder,shahane2023semi}. 
% As we see from \cref{eq:rbfpde}, we tackle this problem partly by employing an enhanced isotropic node arrangement strategy initiated in \cite{zamolo2019solution}. 
Our implementation accounts for all three major boundary conditions in the literature by careful (re)ordering of the nodes: first the $N_i$ internal nodes, then $N_d$ Dirichlet nodes, then $N_n$ Neumann nodes, and finally $N_r$ Robin nodes. We refer the reader to our code \texttt{$\mathbb{U}$pdec} \cite{nzoyem2023updec} for details.

% The optimal control tasks under consideration in this work is that of finding a $c* \in \mathcal{C}$ that minimises a cost objective functional $\mathcal{J}:\mathcal{C} \rightarrow \mathbb{R}$. 
In this work, we consider stationary limits of PDEs with no time dependence. Also, our control functions $c:\mathcal{C}\rightarrow \mathbb{R}$ are only applied to part of the boundaries. To provide a unified framework suitable for all experiments we conduct, we generalise \cref{eq:pde1,eq:pde2,eq:pde3,eq:pde4} by accounting for vector fields with
\begin{subequations}    \label{eq:pde}
\begin{empheq} [left=\empheqlbrace] {align}
    \mathcal{F}(\mathbf{u}) &= 0 &&\text{ in } \Omega \\
    \mathcal{B}(\mathbf{u},c) &= 0, &&\text{ on } \partial \Omega
\end{empheq}
\end{subequations}
where $\mathcal{F}$ and $\mathcal{B}$ are respectively the PDE and boundary residuals. The state $\mathbf{u}$ is fully dependent on the control $c$, and the constrained optimisation problem is to find
\begin{align}
    c^* = \argmin_{c}{\mathcal{J}(c)} \quad \text{ subject to \cref{eq:pde} }.
\end{align}

\subsection{Direct-Adjoint Looping}

The use of continuous adjoint variables for optimisation under equality constraints has its origins in Lagrange multipliers \cite{lagrange1853mecanique}. Centuries of development culminating in the Karush-Kuhn-Tucker conditions---to account for inequality constraints---\cite{kuhn2013nonlinear} and Pontryagin's Maximum Principle in calculus of variations \cite{pontryagin1962maximum} have set the groundwork  for optimal control theory and computational experimentation via adjoint-based sensitivity analysis.

To derive the adjoint equations, one may use calculus of variations to consider $\varepsilon >0 $ and a small perturbation $h \in \mathcal{C}$, then define the directional derivative
\begin{align*}
D\mathcal{J}(c)\cdot h &= \lim_{\varepsilon \rightarrow 0}{ \frac{\mathcal{J}(c+\varepsilon h) - \mathcal{J}(c)}{\varepsilon }}, \\
&= \frac{\md\mathcal{J}(c+\varepsilon h)}{\md\varepsilon} \bigg\rvert_{\varepsilon = 0},
\end{align*}
whose expansion should display the state sensitivity w.r.t. the control
$$\mathbf{u}'(h) = \lim_{\varepsilon \rightarrow 0}{ \frac{\mathbf{u}_{c+\varepsilon h} - \mathbf{u}_c}{\varepsilon }},$$
where the explicit notation $\mathbf{u}_c$ indicates the solution to \cref{eq:pde} given a control $c$. 
% This highlights the strong connection between $\mathbf{u}$ and $c$. 
Linearity and continuity of $\mathbf{u}'$ w.r.t. $h$ may then be established by deriving a separate PDE based on \cref{eq:pde};
% \begin{align} \label{eq:pdeprime}
%     \mathcal{F'}(\mathbf{u'}) &= 0 \text{ in } \Omega \\
%     \mathcal{B'}(\mathbf{u'},\mathbf{c}) &= 0 \text{ on } \partial \Omega
% \end{align}
both sides of which are multiplied by an adjoint state $\mathbf{\lambda}$ (independent of $c$). Integration by parts and careful consideration of boundary conditions are then used to isolate $\mathbf{u}'$ from the differential operators, leading to an adjoint PDE for $\mathbf{\lambda}$
\begin{subequations}    \label{eq:pdeprime}
\begin{empheq} [left=\empheqlbrace] {align}
    \mathcal{F_a}(\mathbf{u}, \mathbf{\lambda}) &= 0 &&\text{ in } \Omega \\
    \mathcal{B_a}(\mathbf{u},c, \mathbf{\lambda}) &= 0, &&\text{ on } \partial \Omega
\end{empheq}
\end{subequations}
after which $\frac{\md\mathcal{J}}{\md c} = \nabla \mathcal{J}$ can be expressed as a function of $\mathbf{u}, \mathbf{\lambda}$ and $c$.
 
As showcased in \cref{fig:DALmethod}, DAL is an \emph{optimise-then-discretise} (OTD) scheme that initialises $c$, then solves \cref{eq:pde}, then \cref{eq:pdeprime}, then evaluates $\nabla \mathcal{J}$, and finally updates $c$ via gradient descent. Details of DAL derivations for the Laplace and Navier-Stokes PDEs using Euler-Lagrange equations can be found in \cite{mowlavi2023optimal}.

\subsection{Physics-Informed Neural Networks}

Similar to RBFs, Multilayer Feed Forward Networks are universal approximators \cite{hornik1989multilayer}. 
% whose success this past decade in representing complex functions is undeniably tied to the abundance of data \cite{halevy2009unreasonable}.
They are leveraged by PINNs \cite{raissi2019physics} to approximate solutions to problems involving PDEs (see \cref{fig:PINNmethod}). PINNs are learnt by enforcing the governing equations as soft constraints at points in the domain and its boundary. The core idea is to include these constraints as part of the loss function in \cref{eq:pinn}
\begin{align} \label{eq:pinn}
    \mathcal{L}(\mathbf{u}_{\theta}, c_{\theta}) = \underbrace{\mathcal{F}(\mathbf{u}_{\theta}) + \mathcal{B}(\mathbf{u}_{\theta}, c_{\theta})}_{\mathcal{L}_{\mathcal{F}/\mathcal{B}}(\mathbf{u}_{\theta}, c_{\theta})}
     + \, \textcolor{red}{\omega}\mathcal{J}(c_{\theta}), 
\end{align}
where the subscripts on $\mathbf{u}$ and $c$ indicate the dependence to Neural Network weights $\theta$. Adding a loss term for labelled data is optional, and the use of disconnected nodes in a point cloud makes the method mesh-free, just like RBFs. PINNs have been used in a wide range of application areas, and their efficiency for forward problems is the subject of considerable research interest.

The usability of PINNs for optimal control, however, remains much less explored, something we address with this work. An impressive study of PINNs for such applications was recently conducted in \cite{mowlavi2023optimal}. Their strategy involved adding to the loss function the cost objective $\mathcal{J}$ weighted by an adjustable coefficient $\omega$. Given the inherent difficulty is solving such a multi-objective optimisation problem, the authors of \cite{mowlavi2023optimal} presented a \emph{two-step line search} strategy for $\omega$, which we reproduce. Before the search starts, a Multilayer Perceptron (MLP) is trained to solve the forward PDE with a prescribed control term. This helps identify the set of hyperparameters suitable for the problem at hand: namely, a learning rate schedule and an architecture for the solution network. Then the two-step line search is as follows:
\begin{enumerate}
    \item A range\footnote{Our recommendation is to start with 1 and explore both directions with positive and negative powers of 10. See examples of ranges in \cref{sec.results}.} of coefficients $\omega$ are used to learn potential optimal controls. Each weight corresponds to a different pair of $(\mathbf{u}_{\theta}, c_{\theta})$ solution-control networks; all trained with the same architecture and hyperparameters preliminarily determined. The weights of $\mathbf{u}_{\theta}$ and $c_{\theta}$ are updated in an alternating manner to minimise the multi-objective loss function $\mathcal{L}_{\mathcal{F}/\mathcal{B}} + \omega\mathcal{J}$.
    \item Since fitting the PDE is an imperative, new solution networks $\mathbf{u}_{\theta}'$ are retrained for each $\omega$. This is done using the control networks saved from step 1. The loss function in this step does not include the cost objective $\mathcal{J}$. After training, the pair $(\mathbf{u}_{\theta}'^*, c^*_{\theta})$ that results in the lowest cost objective is chosen as the optimal solution.\footnote{We find it crucial to train $\mathbf{u}_{\theta}'$ at least until it matches $c_{\theta}$ on the appropriate boundaries.}
\end{enumerate}

\subsection{Differentiable Programming}

The rise of Deep Learning was accompanied by a substantial development in tools to efficiently calculate derivatives \cite{griewank2008evaluating,jax2018github}. For instance, with PINNs, network weights are updated via backpropagation \cite{werbos1974beyond}, which is a special case of automatic differentiation (AD) \cite{baydin2018automatic}. In contrast to Finite Differences which suffers from truncation and round-off errors, AD returns the exact gradients. It relies on the chain rule of differentiation applied to elementary built-in or custom-made units to compute exact values without ever analytically deriving the corresponding symbolic expressions.

\begin{figure}
     \centering
     \begin{subfigure}[b]{0.479\linewidth}
         \centering
         \frame{\includegraphics[width=\textwidth]{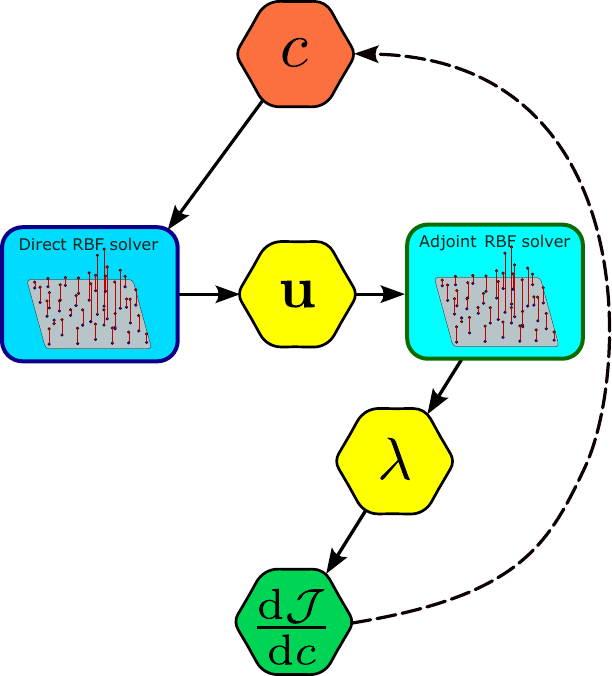}}
         \caption{DAL}
         \label{fig:DALmethod}
     \end{subfigure}
     % \hfill
     \begin{subfigure}[b]{0.47\linewidth}
         \centering
         \frame{\includegraphics[width=\textwidth]{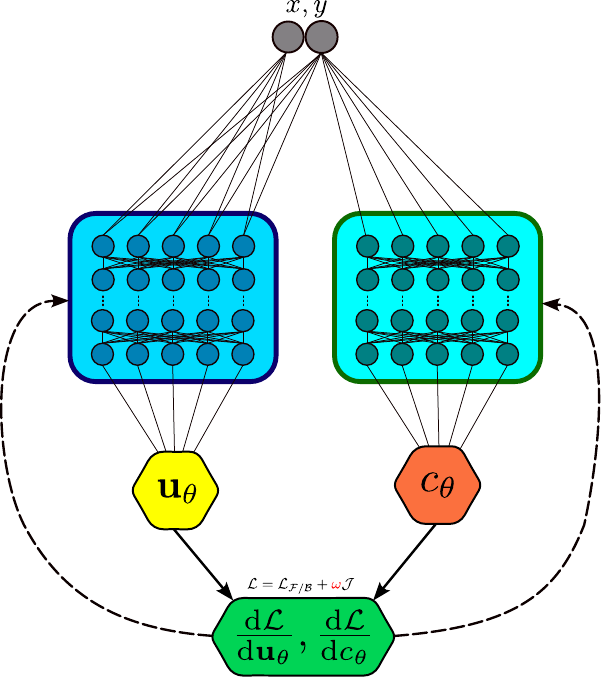}}
         \caption{PINN}
         \label{fig:PINNmethod}
     \end{subfigure}
     % \hfill
     \begin{subfigure}[b]{0.96\linewidth}
         \centering
         \frame{\includegraphics[width=\textwidth]{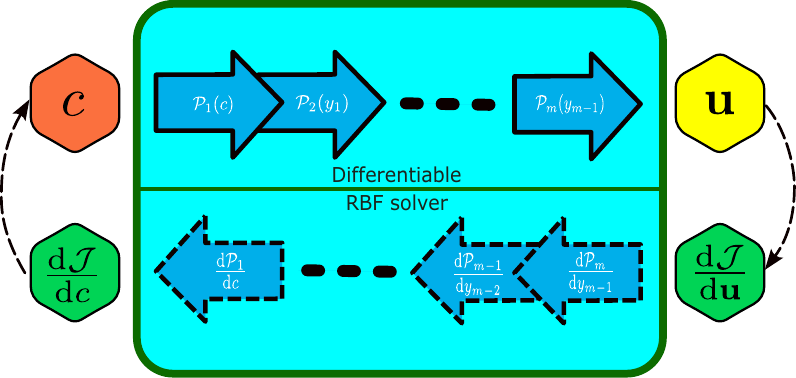}}
         \caption{DP}
         \label{fig:DPmethods}
     \end{subfigure}
        \caption{Illustration of a) DAL as it evaluates $\nabla \mathcal{J}$ by simulating two distinct PDEs. b) PINN fits a Neural Network by enforcing soft PDE constraints, which $\mathcal{J}$ is part of. c) DP combines the fitting strategy of PINNs with the strict adherence to first principles of DAL; gradients travel backwards through the solver before influencing $c$.}
        \label{fig:methods}
\end{figure}

In recent years, the tools that made AD so successful for Deep Learning have been leveraged for simulation in the DTO paradigm of differentiable programming (DP), also known as differentiable physics, differentiable simulation, or differentiable modelling. Showcase examples of applications of AD outside or in conjunction with Neural Networks include \cite{holl2020learning,bezgin2023jax,nimier2019mitsuba,schoenholz2019jax,shen2023differentiable}. The same concept permeates through all those examples: they are a succession of elementary operators $\{\mathcal{P}\}_{i=1,\ldots,m}$ whose derivatives w.r.t. to their inputs can be evaluated (see \cref{fig:DPmethods}). That is how $\nabla \mathcal{J}$ is ultimately obtained.

Our RBF software \texttt{$\mathbb{U}$pdec} \cite{nzoyem2023updec} consists of such operations, which make the solver end-to-end differentiable. We use JAX \cite{jax2018github} and its reverse-mode implementation of AD called \texttt{grad} to compute $\nabla \mathcal{J}$. In addition, the \texttt{grad} transformation is used to define the differential operator $\mathcal{D}$ described in section \ref{subsec:rbf}, thus giving users the liberty to effortlessly choose or design new functions $\phi$ (see \cref{eq:rbffield}). Besides \texttt{grad}, JAX is particularly suited for this task because of its just-in-time compilation \texttt{jit} and batched (or vectorisation) \texttt{vmap} transformations which improve solver runtimes.

% \begin{figure*}
%   \includegraphics[width=\textwidth, height=5cm]{figures/aproaches.png}
%   \caption{a) DAL evalues $\nabla J$ by simulating two distinc PDEs based on first-principles. b) PINN fits a Neural Network by enforcing soft PDE constraints at an arbitrary number of points. c) DP combines the fitting strategy of PINNs with the strict adherence to first principles of DAL.}
% \end{figure*}

\section{Results}
\label{sec.results}

The accuracy and convergence rate of any RBF-based simulation depend heavily on the properties of the basis function $\phi$. The merits and setbacks of several choices are discussed in the literature \cite{shahane2023semi}: multiquadrics, Gaussians, thin plate splines, etc; most of which are parametrised by a shape parameter. To avoid tuning such parameter, we opted for the polyharmonic cubic spline $\phi: r \rightarrow r^3$, which when augmented with polynomials of maximum degree\footnote{In 2D i.e., $d=2$, setting $n=1$ results in appending $M = \binom{n+d}{n}$ = 3 polynomials.} $n=1 $, provided a robust and performant tool, capable of approximating even non-linear PDEs such as the Navier-Stokes equations.  

For all our DAL, PINN, and DP experiments, we used the Adam optimiser \cite{gal2016dropout} to perform gradient descent. While firmly established for Neural Network optimisation, the use of Adam with DAL or DP is less common. In this study, Adam helped increase robustness to noisy gradients at boundaries due to the Runge phenomenon \cite{fornberg2007runge}. To guide our schemes towards faster and more accurate convergence to a minimiser, the initial learning rate was divided by 10 after half the iterations or epochs, and again by 10 at 75\% completion.

\subsection{The Laplace Problem}

Consider the Laplace equation \cite{mowlavi2023optimal} in the unit square $\Omega$ with Dirichlet boundary conditions 
\begin{subequations}    \label{eq:laplaceproblem}
\begin{empheq} [left=\empheqlbrace] {align}
    \frac{\partial^2 u}{\partial x^2} + \frac{\partial^2 u}{\partial y^2} &= 0    && \text{in } \Omega  \label{eq:laplacein} \\
    u(x,1) &= c(x) \\ 
    u(x,0) &= \sin \pi x   && \text{on } \partial \Omega\\
    u(0,y) = u(1,y) &= 0,
\end{empheq}
\end{subequations}
where $c$ is the control potential applied to the top wall. We seek to solve the convex optimal control problem
\begin{align}
    c^* = \argmin_{c} \mathcal{J}(c) \quad \text{subject to \cref{eq:laplaceproblem} }, 
\end{align}
with
\begin{align*}
    \mathcal{J}(c) = \int_0^1 \Big\vert \frac{\partial u}{\partial y}(x,1) - \cos \pi x \Big\vert^2 \md x. 
\end{align*}
This problem has an analytical minimiser given by
\begin{multline*}
    c^*(x) = \sech(2\pi)\sin(2\pi x) + \frac{1}{2\pi}\tanh(2\pi)\cos(2\pi x),
\end{multline*}
corresponding to the state solution
\begin{multline*}
    u^*(x,y) = \frac{1}{2}\sech(2\pi)\sin(2\pi x) \left( \e^{2\pi (y-1)} + \e^{2\pi (1-y)} \right)\\ + \frac{1}{4\pi} \sech(2\pi)\cos(2\pi x) \Big( \e^{2\pi y} - \e^{-2\pi y} \Big).
\end{multline*}

For the DAL and DP techniques, the PDE \eqref{eq:laplaceproblem} was solved on a regular 100 $\times$ 100 grid, which resulted in better conditioned collocation matrices compared with a scattered point cloud of the same size. $\nabla \mathcal{J}$ was then evaluated to iteratively update $c$, initially set to identically $0$. The initial learning rates for DAL and DP was $10^{-2}$, and $10^{-3}$ for the PINN\footnote{We present hyperparameters for the first and most important step of the line search strategy, which often required values different from step 2.}.

With the PINN, training as seen in \cref{fig:pinnlosses1,fig:pinnlosses2,fig:pinnchoice} was done on a scattered cloud, while testing was performed on the same 100$\times$100 regular grid as for DAL and DP. This regularised the PINN and improved generalisation. The preliminary step to the line search recommended using a MLP with 3 hidden layers of 30 neurons each, an architecture which provided good balance between expressiveness, overfitting, and computational efficiency. Each layer was equipped with an infinitely differentiable \texttt{tanh} activation function. Like in \cite{mowlavi2023optimal}, we tried 11 values of $\omega$ (from $10^{-3}$ to $10^7$) and analysed their effect on the total loss function \cref{eq:pinn}. We found that $\omega^*=10^{-1}$ gives rise to the most balanced solution for this problem. Results highlighting the unmatched performance of DP are reported in \cref{fig:laplaceresults}; and a summary of the main hyperparameters used throughout the experiment is presented in \cref{tab:laplace}.

\begin{table}[h]
    \centering
    \begin{tabular}{l c c c} 
        \toprule
        \tabhead{Hyperparameter}& \tabhead{DAL} & \tabhead{PINN} & \tabhead{DP} \\
        \midrule
        Init. learning rate &  $10^{-2}$ & $10^{-3}$ & $10^{-2}$ \\
        Batch size &  - & 1000 & - \\
        Network architecture &  - & 3$\times$30 & - \\
        Epochs &  - & 20k & - \\
        Iterations &  500 & - & 500 \\
        Point cloud size &  $10^4$ & $10^4$ & $10^4$ \\
        Max. polynomial degree $n$ &  1 & - & 1 \\
        \bottomrule
    \end{tabular} \vspace*{2mm}
    \caption{Hyperparameter summary for the Laplace problem implemented in a Python \cite{van2007python} environment equipped with JAX \cite{jax2018github} and \texttt{$\mathbb{U}$pdec} \cite{nzoyem2023updec}. The learning rates follow a piece-wise constant schedule described early in \cref{sec.results}. The network architecture indicates the number of hidden layers and neurons per layer in the MLP. The hyphen symbol "-" stands for hyperparameters not applicable (NA) to the method under consideration.}.
    \label{tab:laplace}
\end{table}

\subsection{The Navier-Stokes Problem}

The continued relevance of fluid flow between parallel plates is clear from extensive studies in the literature \cite{raissi2018hidden,attia1996mhd}. As such, it represents an excellent case study for our methods.
\begin{figure}
     \centering
     \begin{subfigure}[b]{0.412\linewidth}
         \centering
         \includegraphics[clip,width=\textwidth, trim = {0 0.5cm 0 0.2cm}, clip]{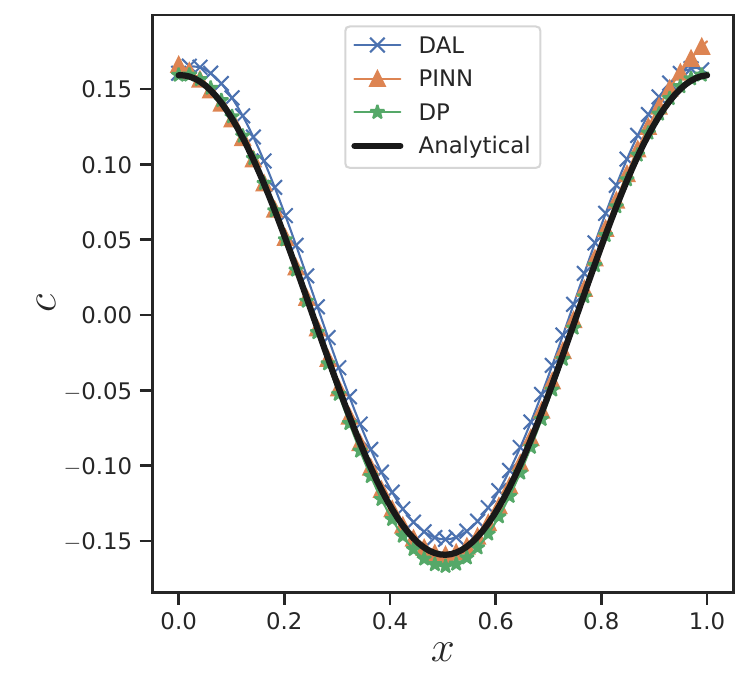}
         \caption{Top wall controls}
         \label{subfig:laplacecontrols}
        \vspace*{2mm}
     \end{subfigure}
     % \hfill
     \begin{subfigure}[b]{0.4\linewidth}
         \centering
         \includegraphics[clip,width=\textwidth, trim = {0 0.35cm 0 0.2cm}, clip]{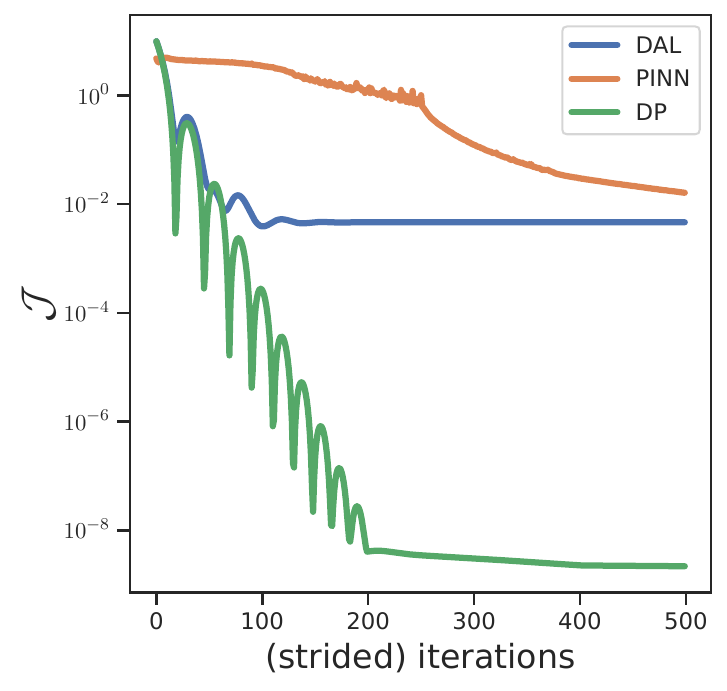}
         \caption{Cost objectives}
         \label{subfig:laplacecosts}
        \vspace*{2mm}
     \end{subfigure}
     \hfill
     \begin{subfigure}[b]{0.3\linewidth}
         \centering
         \includegraphics[width=\textwidth]{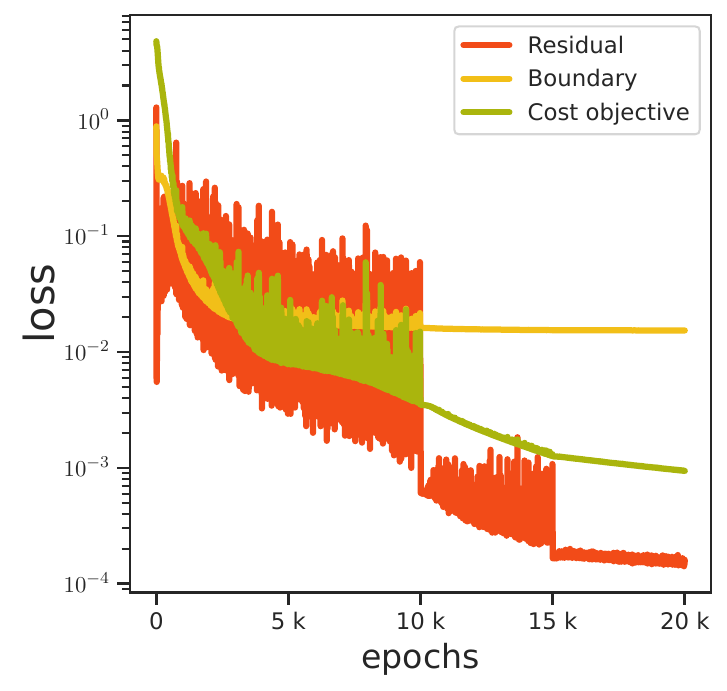}
         \caption{\tiny Line search step 1 ($\omega=0.1$)}
         \label{fig:pinnlosses1}
        \vspace*{2mm}
     \end{subfigure}
     \hfill
     \begin{subfigure}[b]{0.3\linewidth}
         \centering
         \includegraphics[width=\textwidth]{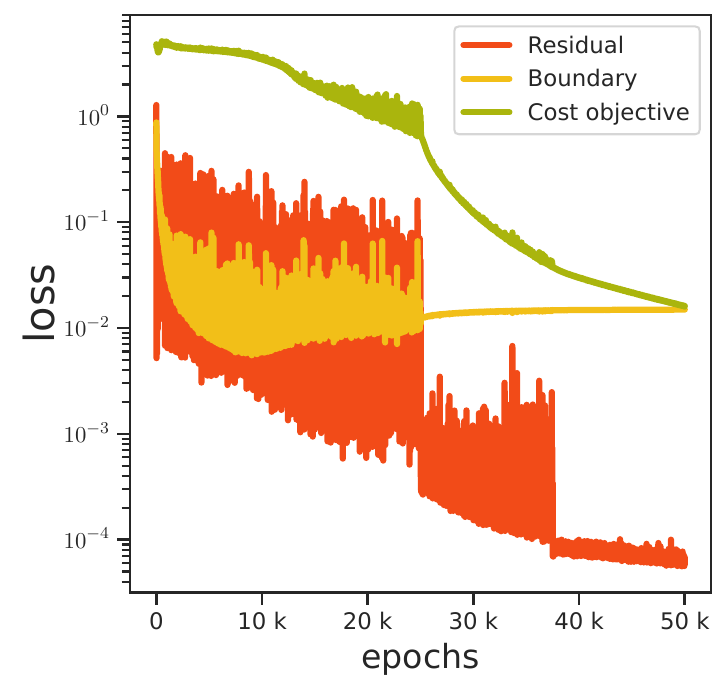}
         \caption{\tiny Line search step 2 ($\omega=0.1$)}
         \label{fig:pinnlosses2}
        \vspace*{2mm}
     \end{subfigure}
     \hfill
     \begin{subfigure}[b]{0.3\linewidth}
         \centering
         \includegraphics[width=\textwidth]{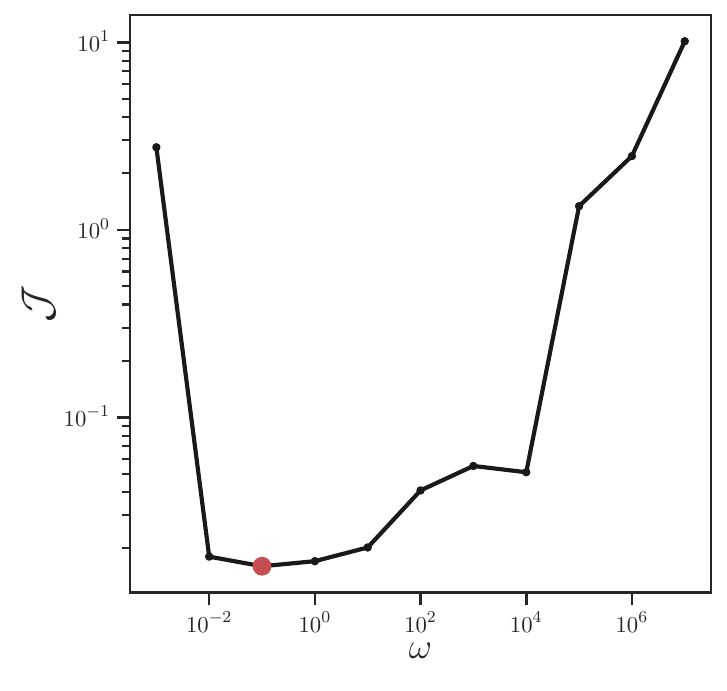}
         \caption{\tiny PINN weight choice}
         \label{fig:pinnchoice}
        \vspace*{2mm}
     \end{subfigure}
     \hfill
     \begin{subfigure}[b]{0.28\linewidth}
         \centering
         \includegraphics[width=\textwidth, trim = {0 1.2cm 0 0.2cm}, clip]{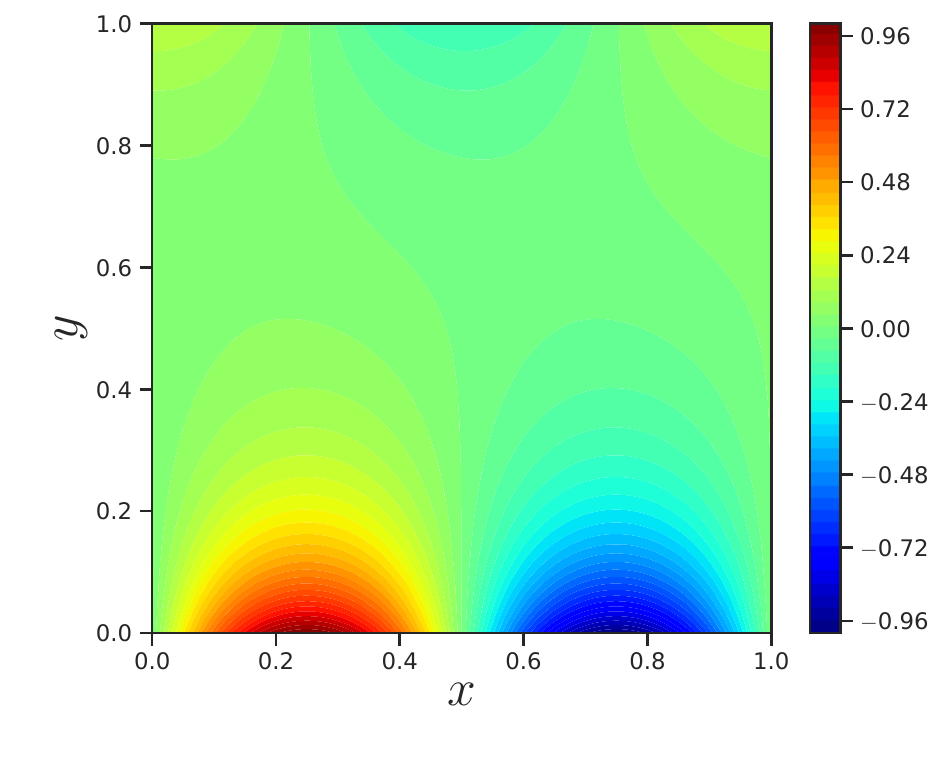}
         \caption{Exact sol.}
         \label{subfig:laplaceexact}
     \end{subfigure}
     \hfill
     \begin{subfigure}[b]{0.706\linewidth}
         \centering
         \includegraphics[clip,width=\textwidth, trim = {0 1.2cm 0 0.2cm}, clip]{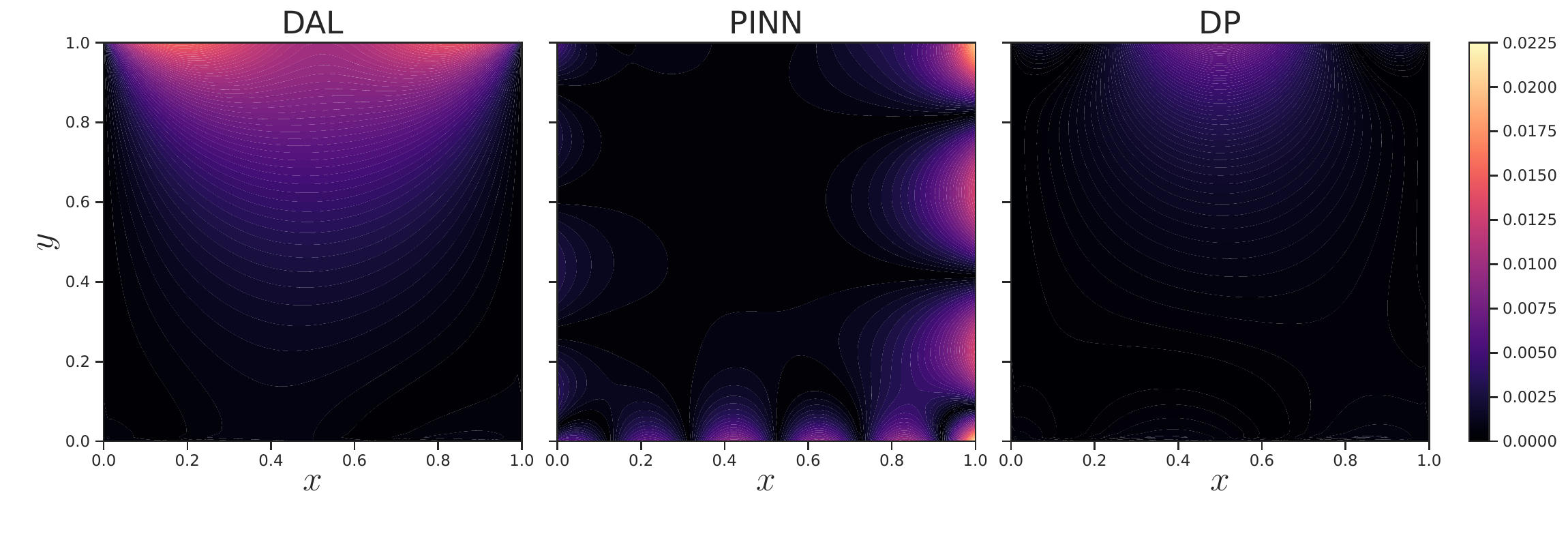}
         \caption{Absolute errors}
         \label{subfig:laplaceerrors}
     \end{subfigure}
        \caption{Results for the Laplace control problem highlighting how well DP is able to minimise the cost objective $\mathcal{J}$ relative to DAL and PINN\protect\footnotemark (see (a) and (b)). In (c), (d), and (e), we showcase the various stages of the PINN's line search strategy for $\omega$. (f) and (g) provide a comparison of the state solutions after optimisation, confirming the superiority of DP through its low absolute error.}
        \label{fig:laplaceresults}
\end{figure}
\footnotetext{The three methods are trained for different number of epochs/iterations as per table \ref{tab:performance}. The term “strided” in \cref{subfig:laplacecosts} is meant to convey that fact.}

% \begin{figure}
%   \includegraphics[width=.5\textwidth]{figures/NavierStokes/channel.png}
%   \caption{Setup for the Navier-Stokes problem. Fluid is perturbed by blowing and suction regions on the boundary. The outlet is a free surface where we want the velocity to be parabolic. Figure reused with permission from \cite{mowlavi2023optimal}.}
%   \label{fig:channel}
% \end{figure}

\begin{figure}
     \centering
     \begin{subfigure}[b]{0.6\linewidth}
          \centering
          \includegraphics[clip,width=\textwidth]{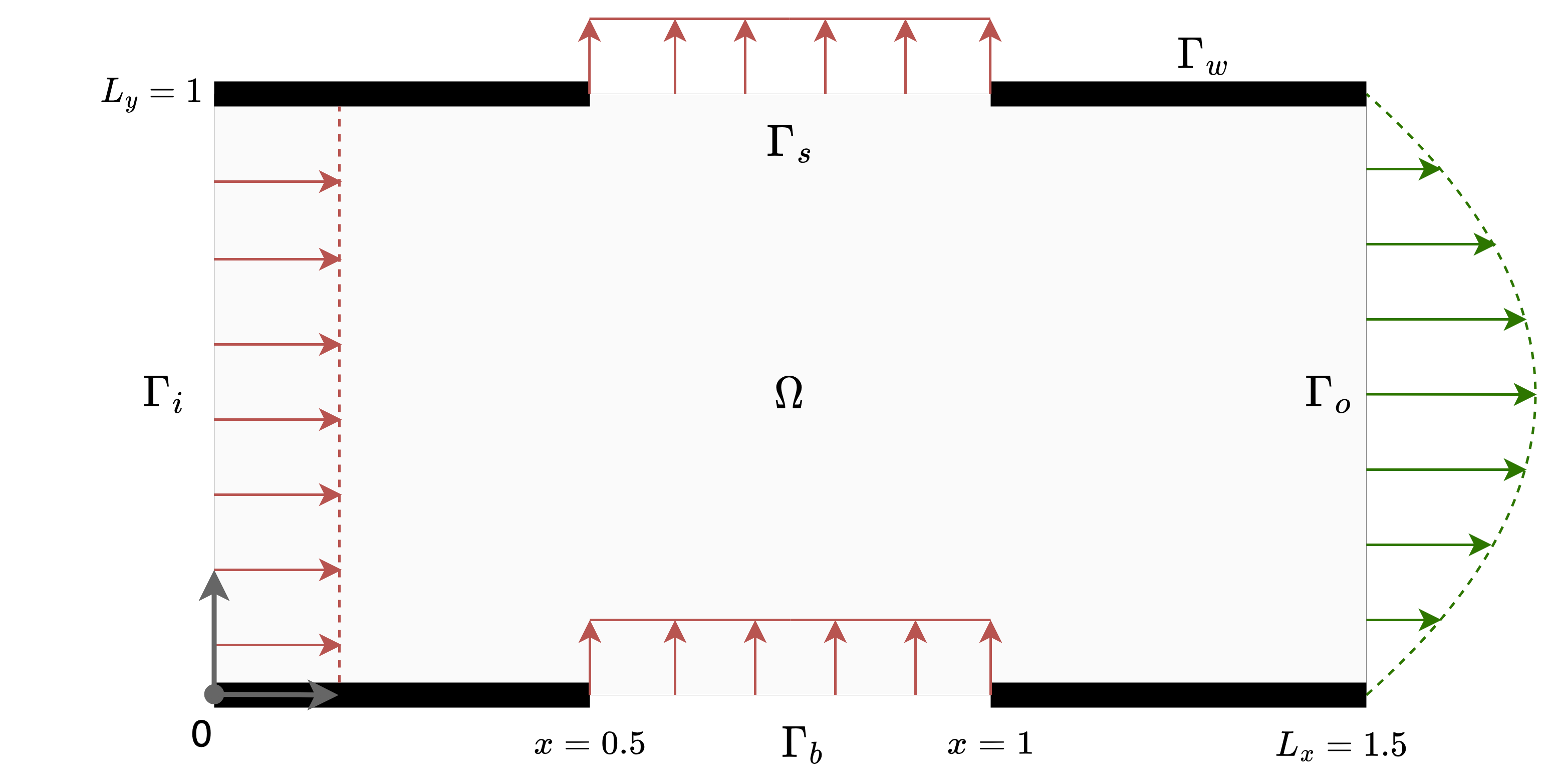}
          % \includesvg[width=\textwidth]{figures/NavierStokes/NavierStokesChannel.svg}
          \caption{Setup}
          \label{fig:channel}
        \vspace*{2mm}
     \end{subfigure}
     \begin{subfigure}[b]{0.3\linewidth}
         \centering
         \includegraphics[clip,width=\textwidth]{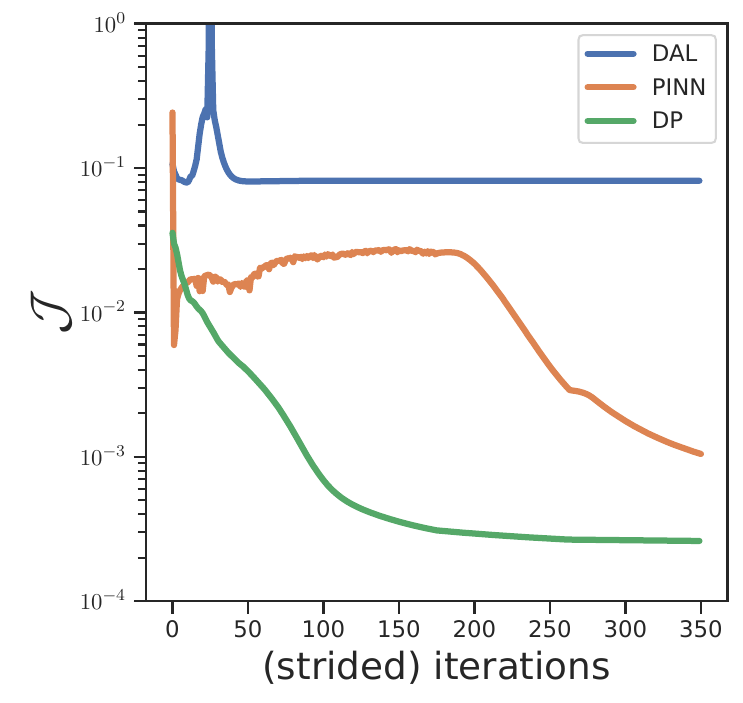}
         \caption{Cost objectives}
         \label{fig:nscosts}
        \vspace*{2mm}
     \end{subfigure}
     % \begin{subfigure}[b]{0.45\linewidth}
     %     \centering
     %     \includegraphics[width=\textwidth]{figures/NavierStokes/pinn_losses_step_1.pdf}
     %     \caption{PINN step 1 ($\omega=1$)}
     %     \label{fig:nspinnlosses}
     % \end{subfigure}
     % % \hfill
     % \begin{subfigure}[b]{0.45\linewidth}
     %     \centering
     %     \includegraphics[width=\textwidth]{figures/NavierStokes/pinn_choice.pdf}
     %     \caption{PINN weight choice}
     %     \label{fig:nspinnschoice}
     % \end{subfigure}
     % \hfill
     \begin{subfigure}[b]{0.45\linewidth}
         \centering
         \includegraphics[clip,width=\textwidth]{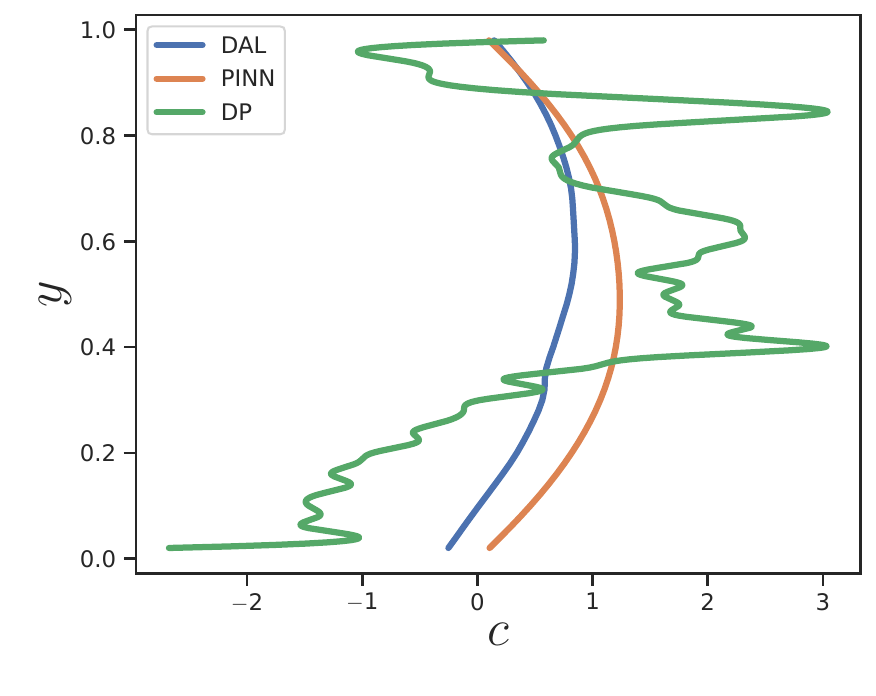}
         \caption{Inflow controls}
         \label{fig:nscontrols}
     \end{subfigure}
     % \hfill
     \begin{subfigure}[b]{0.446\linewidth}
         \centering
         \includegraphics[clip,width=\textwidth]{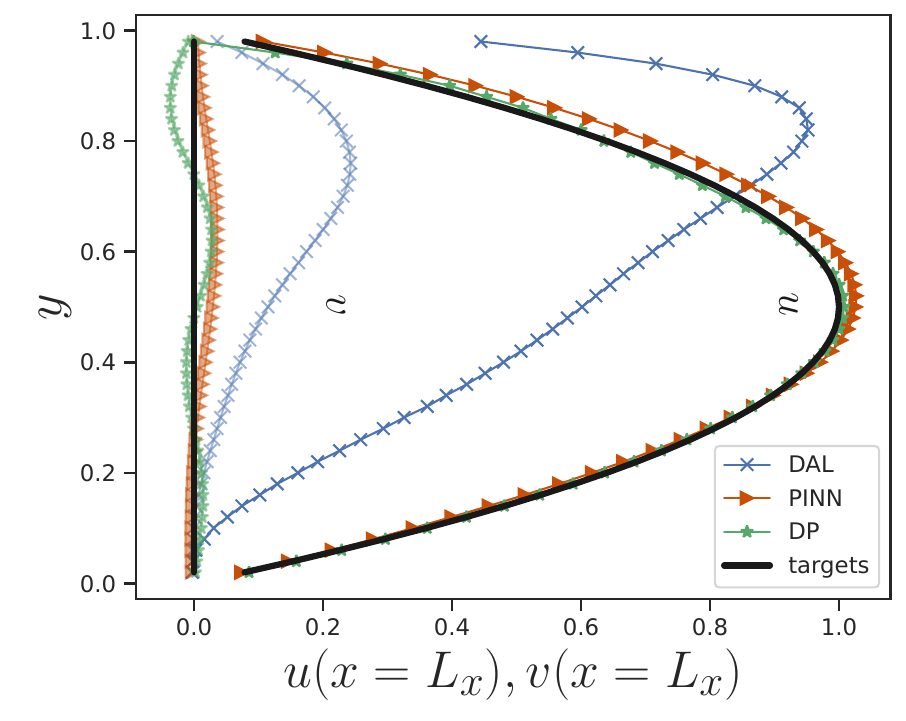}
         \caption{Outflow velocities}
         \label{fig:nsoutvels}
     \end{subfigure}
        \caption{Setup adapted from \cite{mowlavi2023optimal} and results for the Navier-Stokes optimal control problem. (d) shows how DP and PINN succeed in achieving a near-parabolic profile at the outlet. However, from \cref{fig:teaser}, we see that PINN achieves good control at the expense of first principles.}
        \label{fig:nsresults}
\end{figure}

The setup and the domain of interest in \cite{mowlavi2023optimal} are illustrated in \cref{fig:channel}. Given some perturbations due to blowing at $\Gamma_b$ and suction at $\Gamma_s$, what inflow velocity at $\Gamma_i$ would cause a parabolic profile at the outlet $\Gamma_o$? The flow in $\Omega$ is governed by the stationary incompressible Navier-Stokes equations in their dimensionless form 
\begin{subequations} \label{eq:nsproblem}
\begin{empheq} [left=\empheqlbrace] {align}
    (\mathbf{u} \cdot \nabla) \mathbf{u} &= -\nabla p + \frac{1}{Re} \nabla^2 \mathbf{u} &&\text{ in } \Omega \\
    \nabla \cdot \mathbf{u} &= 0, &&\text{ in } \Omega
\end{empheq}
\end{subequations}
where $\mathbf{u}=(u,v)$. The (controlled) boundary conditions are
\begin{subequations} \label{eq:nsbcproblem} 
\begin{empheq} [left=\empheqlbrace] {align}
    \mathbf{u} &= (c(y),0) &&\text{ on } \Gamma_i \\
     \mathbf{u} &= (0,0.3) &&\text{ on } \Gamma_b \\
     \mathbf{u} &= (0, 0.3) &&\text{ on } \Gamma_s \\
     \mathbf{u} &= (0,0) &&\text{ on } \Gamma_w \\
     (\mathbf{n} \cdot \nabla)\mathbf{u} &= (0,0)  &&\text{ on } \Gamma_o \\
     (\mathbf{n} \cdot \nabla)p &= 0  &&\text{ on } \Gamma_i \cup \Gamma_b \cup \Gamma_s \cup \Gamma_w \\
     p &= 0.  &&\text{ on } \Gamma_o
\end{empheq}
\end{subequations}
We want to find
\begin{align}
    c^* = \argmin_{c} \mathcal{J}(c) \quad \text{subject to  \cref{eq:nsproblem,eq:nsbcproblem}},
\end{align}
with
\begin{align*}
    \mathcal{J}(c) = \frac{1}{2}\int_0^{L_y} \Big( \big\vert u(L_x,y) - \frac{4}{L_y^2}y(1-y) \big\vert^2 + \big\vert v(L_x,y) \big\vert^2 \Big) \md y. 
\end{align*}

Upon deriving the adjoint equations, we decoupled the (adjoint) velocity components and the (adjoint) pressure. We employed a Chorin-inspired projection approach \cite{chorin1967numerical} to iteratively bring the fields to steady states \cite{zamolo2019solution}. We set the number of refinements $k=3$ for DAL and $k=10$ for DP, both with initial learning rates $10^{-1}$. We set the Reynolds number to $Re=100$, and the initial guess for the inflow velocity to $4y(1-y)/L_y^2$. Given the complexity of the domain and the benefits of mesh refinement near free surfaces \cite{zamolo2019solution}, we meshed the domain with GMSH \cite{geuzaine2009gmsh}, from which we extracted 1385 scattered and disconnected nodes, used for all three methods (see \cref{tab:navierstokes}). 

In the PINN's loss function, we included all Dirichlet and homogeneous Neumann boundary penalty terms for the velocity as suggested in \cite{mowlavi2023optimal}. As for the pressure, only the Dirichlet boundary conditions at the outlet were enforced since it made no difference to the preliminary calibration of the PINN architecture to include Neumann conditions. During this preliminary step, we found that a MLP architecture with 5 fully connected hidden layers of 50 neurons each was well suited for this problem, and it offered a limited computational cost. As seen in \cref{tab:navierstokes}, the initial learning rate was set to $10^{-3}$. The line search strategy explored 9 values for $\omega$ from $10^{-3}$ to $10^{5}$, settling on $\omega^*=1$.

\begin{table}[h]
    \centering
    \begin{tabular}{l c c c} 
        \toprule
        \tabhead{Hyperparameter}& \tabhead{DAL} & \tabhead{PINN} & \tabhead{DP} \\
        \midrule
        Init. learning rate &  $10^{-1}$ & $10^{-3}$ & $10^{-1}$ \\
        Network architecture &  - & 5$\times$50 & - \\
        Epochs &  - & 100k & - \\
        Iterations &  350 & - & 350 \\
        Refinements $k$ &  3 & - & 10 \\
        Point cloud size &  1385 & 1385 & 1385 \\
        Max. polynomial degree $n$ &  1 & - & 1 \\
        \bottomrule
    \end{tabular} \vspace*{2mm}
    \caption{Hyperparameter summary for the Navier-Stokes problem. Like in \cref{tab:laplace}, the hyphen "-" stands for not applicable (NA).}.
    \label{tab:navierstokes}
\end{table}

% The results are reported at \cref{fig:nsresults}.
The DP inflow solution as seen in \cref{fig:nscontrols} yields the best fitting outflow velocity in \cref{fig:nsoutvels}.
The DAL fails to capture the solution due to RBF-related inaccuracies when computing the gradients of $\mathbf{u}$, needed for resolving the adjoint advection operator. We found that this problem is lessened with a reduced $Re=10$ which led to better solutions with DAL. The DP and the PINN on the other hand, succeed in capturing minima, with the cost objectives getting as low as $2.61\times 10^{-4}$ and $1.04\times 10^{-3}$, respectively. 

\section{Discussion}
\label{sec.discusion}

In the context of RBFs, the DAL approach was shown to perform well on the Laplace optimal control problem, although it compared poorly to the other methods (see \cref{subfig:laplacecosts}). Here, DAL converged despite the gradients rising to very large values. The Adam optimiser which alleviated this issue was ultimately unable to repeat the same for the Navier-Stokes problem (see \cref{fig:nscosts}). This is common in OTD approaches, where round-off and numerical errors are often encountered \cite{kidger2022neural}. In short, with DAL, numerical errors linked to the derivation of continuous adjoint equations should be handled with care.

A well implemented PINN along with a two-step line search strategy is a powerful tool for optimal control. The PINN was able to find appropriate controls for the Laplace and Navier-Stokes problems (see \cref{subfig:laplacecontrols,fig:nscontrols}). Although setting up and training a PINN is perhaps more costly compared to DAL (which is very problem-specific), its performance is good. These observations align with those of \cite{mowlavi2023optimal}
% whose seminal work inspired this paper
. On the flip side, the PINN showed little generalisation capability, especially with regard to $Re$. We experimented with values from $Re=10$ to $Re=100$ and found that more laminar flows required significantly more epochs to train to satisfaction\footnote{The results concerning the generalisation of PINNs with regard to $Re$ are not reported in this work.}. The idea of repeating the line search strategy for each new set of parameters diminishes the appeal of PINNs. Furthermore, PINNs require a considerable amount of data and a substantial training time \cite{shen2023differentiable} as evident in \cref{tab:performance}.

The most well-rounded approach is undeniably the DP. It provides extremely low cost objectives for both Laplace and Navier-Stokes problems (see \cref{subfig:laplacecosts,fig:nscosts}). In addition to it being relatively easy to set up, its \emph{discretise-then-optimise} (DTO) approach is able to overcome the RBF-related inaccuracies that plague the DAL. That said, DP as conceived in this study can be memory inefficient due to storage and optimisation of a computational graph \cite{kidger2022neural,jax2018github}. For the Navier-Stokes problem, the computational complexity scales super-linearly with the number of refinement steps $k$ needed to account for the non-linear advection operator. As recommended in similar studies \cite{ma2021comparison,kidger2022neural}, if $k$ is small then DP should be prioritised, especially given that its gradients are the gold standard in optimisation\footnote{Although it should be noted that classical Finite Differences was efficient in providing accurate gradients for our Navier-Stokes problem at a reduced memory cost.} \cite{kidger2022neural,shen2023differentiable}.

We also note from \cref{fig:nscontrols} that the DP control is considerably less smooth than the other two.
This could be resolved by increasing the learning rate and performing less gradient descent iterations; or by penalising the control's variations by adding the integral term $\int_0^{L_y} \vert\nabla u (L_x,y)\vert^2 \md y$ to the cost objective.
We refrained from doing the latter since it prevents a fair comparison between the methods.

\begin{table}
    \centering
    \begin{tabular}{c l l l l} 
        \toprule
        \multirow{2}{*}{\tabhead{Problem}} & \multirow{2}{*}{\tabhead{Metric}}& \multicolumn{3}{c}{\tabhead{Method}} \\
        {} & {} & DAL & PINN & DP\\
        \midrule
        \multirow{4}{*}{Laplace} & Time (hours) & 3.3 & 7.3* & 1.65 \\
        &Peak mem. (GB) &  33.6 & 5.0  & 20.2  \\
        &Epochs / Iters. &  500 & 20k & 500 \\
        &Final cost $\mathcal{J}$  & 4.6e-3  & 1.6e-2 & 2.2e-9 \\
        \midrule
        \multirow{5}{*}{Navier-Stokes} & Time (hours) & 1.5 & 26.8* & 3.8 \\
        &Peak mem. (GB) & 8.1  & 1.3 &  45.3 \\
        &Epochs / Iters. &  350 & 100k & 350 \\
        &Refinements &  3 & - & 10 \\
        &Final cost $\mathcal{J}$ & 8.2e-2  & 1.0e-3 & 2.6e-4 \\
        \bottomrule 
    \end{tabular}\vspace*{2mm}
    \caption{Performance details for DAL, PINN, and DP, with each method run on the hardware that allowed optimal performance. DAL and DP results were obtained on a AMD Ryzen 9 5950X 16-Core Processor. The PINN was trained on an Nvidia GeForce RTX 3090 (we highlight this with a *). }
    \label{tab:performance}
\end{table}

The computational performance of the methods is implementation-specific. However, we provide in \cref{tab:performance} a comparative summary of the three methods using \texttt{$\mathbb{U}$pdec} \cite{nzoyem2023updec}. It indicates the resources required to ultimately achieve the reported values of the cost objectives $\mathcal{J}$. Although only indicative, we believe that similar benchmarks should always be considered when solving an optimal control problem.

Interpreting \cref{tab:performance}, we see that DP is applicable in most optimal control scenarios; that is until the memory requirements become prohibitively large. In those extreme cases, we believe the PINN serves as a good second choice. Furthermore, the rapidly growing body of research currently focusing on advanced PINN paradigms for inverse problems will likely address most issues we've outlined in this work. The best use case for DAL is when the problem formulation is not complex (either linear elliptic or parabolic PDEs), allowing for a seamless derivation of the adjoint equations, like we've shown with the Laplace problem.

\section{Conclusion}
\label{sec.conclusion}

This paper compared the expressiveness of Deep Neural Networks to that of Radial Basis Functions for mesh-free control of systems governed by PDEs. We find that a combination of ideas from both worlds via differentiable programming (DP) is the most promising approach. For optimal control problems under Laplace and Navier-Stokes constraints, DP compared favourably to Physics-Informed Neural Networks (PINNs) in terms of accuracy and computational performance. It even succeeded where the well-established direct-adjoint looping (DAL) algorithm failed. We developed a flexible JAX-based framework for carrying out a robust and diverse range of experiments, which allowed us to share indicative comparisons on the performance of each method. Future work on this project will aim at improving the stability of the adjoint DAL equations and incorporate time to tackle turbulent flows. This includes exploring alternative mesh-free methods like Smoothed Particle Hydrodynamics (SPH). Moreover, we aim to improve the memory and computational efficiency of DP by massively parallelising the framework.

%% Acknowledgments
\begin{acks}
This work was supported by the UK Research and Innovation grant EP/S022937/1: Interactive Artificial Intelligence and EPSRC programme grant EP/R006768/1: Digital twins for improved dynamic design. The authors would also like to thank Drs. Nestor Demeure, Saviz Mowlavi, and Saleh Nabi for interesting discussions and invaluable advice throughout.
\end{acks}

%% Balance the last page
\balance
%% Bibliography style
\bibliographystyle{bibformats/ACM-Reference-Format}
\bibliography{references}


\begin{thebibliography}{51}

%%% the Web address.

\ifx \showCODEN    \undefined \def \showCODEN     #1{\unskip}     \fi
\ifx \showDOI      \undefined \def \showDOI       #1{#1}\fi
\ifx \showISBNx    \undefined \def \showISBNx     #1{\unskip}     \fi
\ifx \showISBNxiii \undefined \def \showISBNxiii  #1{\unskip}     \fi
\ifx \showISSN     \undefined \def \showISSN      #1{\unskip}     \fi
\ifx \showLCCN     \undefined \def \showLCCN      #1{\unskip}     \fi
\ifx \shownote     \undefined \def \shownote      #1{#1}          \fi
\ifx \showarticletitle \undefined \def \showarticletitle #1{#1}   \fi
\ifx \showURL      \undefined \def \showURL       {\relax}        \fi
% invisible to TeX
\providecommand\bibfield[2]{#2}
\providecommand\bibinfo[2]{#2}
\providecommand\natexlab[1]{#1}
\providecommand\showeprint[2][]{arXiv:#2}

\bibitem[Acosta(1995)]%
        {acosta1995radial}
\bibfield{author}{\bibinfo{person}{Felipe Miguel~Aparicio Acosta}.} \bibinfo{year}{1995}\natexlab{}.
\newblock \showarticletitle{Radial basis function and related models: an overview}.
\newblock \bibinfo{journal}{\emph{Signal Processing}} \bibinfo{volume}{45}, \bibinfo{number}{1} (\bibinfo{year}{1995}), \bibinfo{pages}{37--58}.
\newblock


\bibitem[Allaire(2015)]%
        {allaire2015review}
\bibfield{author}{\bibinfo{person}{Gr{\'e}goire Allaire}.} \bibinfo{year}{2015}\natexlab{}.
\newblock \showarticletitle{A review of adjoint methods for sensitivity analysis, uncertainty quantification and optimization in numerical codes}.
\newblock \bibinfo{journal}{\emph{Ing{\'e}nieurs de l'Automobile}}  \bibinfo{volume}{836} (\bibinfo{year}{2015}), \bibinfo{pages}{33--36}.
\newblock


\bibitem[Allegretto et~al\mbox{.}(2011)]%
        {allegretto2011coexistence}
\bibfield{author}{\bibinfo{person}{Walter Allegretto}, \bibinfo{person}{Genni Fragnelli}, \bibinfo{person}{Paolo Nistri}, {and} \bibinfo{person}{Duccio Papini}.} \bibinfo{year}{2011}\natexlab{}.
\newblock \showarticletitle{Coexistence and optimal control problems for a degenerate predator--prey model}.
\newblock \bibinfo{journal}{\emph{Journal of mathematical analysis and applications}} \bibinfo{volume}{378}, \bibinfo{number}{2} (\bibinfo{year}{2011}), \bibinfo{pages}{528--540}.
\newblock


\bibitem[Attia and Kotb(1996)]%
        {attia1996mhd}
\bibfield{author}{\bibinfo{person}{Hazem~Ali Attia} {and} \bibinfo{person}{NA Kotb}.} \bibinfo{year}{1996}\natexlab{}.
\newblock \showarticletitle{MHD flow between two parallel plates with heat transfer}.
\newblock \bibinfo{journal}{\emph{Acta mechanica}} \bibinfo{volume}{117}, \bibinfo{number}{1-4} (\bibinfo{year}{1996}), \bibinfo{pages}{215--220}.
\newblock


\bibitem[Baydin et~al\mbox{.}(2018)]%
        {baydin2018automatic}
\bibfield{author}{\bibinfo{person}{Atilim~Gunes Baydin}, \bibinfo{person}{Barak~A Pearlmutter}, \bibinfo{person}{Alexey~Andreyevich Radul}, {and} \bibinfo{person}{Jeffrey~Mark Siskind}.} \bibinfo{year}{2018}\natexlab{}.
\newblock \showarticletitle{Automatic differentiation in machine learning: a survey}.
\newblock \bibinfo{journal}{\emph{Journal of Marchine Learning Research}}  \bibinfo{volume}{18} (\bibinfo{year}{2018}), \bibinfo{pages}{1--43}.
\newblock


\bibitem[Beguinet et~al\mbox{.}(2023)]%
        {Beguinet2023}
\bibfield{author}{\bibinfo{person}{Adrien Beguinet}, \bibinfo{person}{Virginie Ehrlacher}, \bibinfo{person}{Roberta Flenghi}, \bibinfo{person}{Maria Fuente}, \bibinfo{person}{Olga Mula}, {and} \bibinfo{person}{Agustin Somacal}.} \bibinfo{year}{2023}\natexlab{}.
\newblock \showarticletitle{Deep learning-based schemes for singularly perturbed convection-diffusion problems}.
\newblock \bibinfo{journal}{\emph{ESAIM: Proceedings and Surveys}}  \bibinfo{volume}{73} (\bibinfo{date}{8} \bibinfo{year}{2023}), \bibinfo{pages}{48--67}.
\newblock
\showISSN{2267-3059}
\urldef\tempurl%
\url{https://doi.org/10.1051/proc/202373048}
\showDOI{\tempurl}


\bibitem[Bezgin et~al\mbox{.}(2023)]%
        {bezgin2023jax}
\bibfield{author}{\bibinfo{person}{Deniz~A Bezgin}, \bibinfo{person}{Aaron~B Buhendwa}, {and} \bibinfo{person}{Nikolaus~A Adams}.} \bibinfo{year}{2023}\natexlab{}.
\newblock \showarticletitle{JAX-Fluids: A fully-differentiable high-order computational fluid dynamics solver for compressible two-phase flows}.
\newblock \bibinfo{journal}{\emph{Computer Physics Communications}}  \bibinfo{volume}{282} (\bibinfo{year}{2023}), \bibinfo{pages}{108527}.
\newblock


\bibitem[Bonnard et~al\mbox{.}(2006)]%
        {bonnard2006mecanique}
\bibfield{author}{\bibinfo{person}{Bernard Bonnard}, \bibinfo{person}{Ludovic Faubourg}, {and} \bibinfo{person}{Emmanuel Tr{\'e}lat}.} \bibinfo{year}{2006}\natexlab{}.
\newblock \bibinfo{booktitle}{\emph{M{\'e}canique c{\'e}leste et contr{\^o}le des v{\'e}hicules spatiaux}}. Vol.~\bibinfo{volume}{51}.
\newblock \bibinfo{publisher}{Springer Science \& Business Media}.
\newblock


\bibitem[Bradbury et~al\mbox{.}(2018)]%
        {jax2018github}
\bibfield{author}{\bibinfo{person}{James Bradbury}, \bibinfo{person}{Roy Frostig}, \bibinfo{person}{Peter Hawkins}, \bibinfo{person}{Matthew~James Johnson}, \bibinfo{person}{Chris Leary}, \bibinfo{person}{Dougal Maclaurin}, \bibinfo{person}{George Necula}, \bibinfo{person}{Adam Paszke}, \bibinfo{person}{Jake Vander{P}las}, \bibinfo{person}{Skye Wanderman-{M}ilne}, {and} \bibinfo{person}{Qiao Zhang}.} \bibinfo{year}{2018}\natexlab{}.
\newblock \bibinfo{booktitle}{\emph{{JAX}: composable transformations of {P}ython+{N}um{P}y programs}}.
\newblock
\urldef\tempurl%
\url{http://github.com/google/jax}
\showURL{%
\tempurl}


\bibitem[Cheney(1966)]%
        {cheney1966introduction}
\bibfield{author}{\bibinfo{person}{E.W. Cheney}.} \bibinfo{year}{1966}\natexlab{}.
\newblock \bibinfo{booktitle}{\emph{Introduction to Approximation Theory}}.
\newblock \bibinfo{publisher}{McGraw-Hill Book Company}.
\newblock
\showISBNx{9780070107571}
\showLCCN{65025916}
\urldef\tempurl%
\url{https://books.google.co.uk/books?id=9uZQAAAAMAAJ}
\showURL{%
\tempurl}


\bibitem[Chorin(1967)]%
        {chorin1967numerical}
\bibfield{author}{\bibinfo{person}{Alexandre~Joel Chorin}.} \bibinfo{year}{1967}\natexlab{}.
\newblock \showarticletitle{The numerical solution of the Navier-Stokes equations for an incompressible fluid}.
\newblock \bibinfo{journal}{\emph{Bull. Amer. Math. Soc.}} \bibinfo{volume}{73}, \bibinfo{number}{6} (\bibinfo{year}{1967}), \bibinfo{pages}{928--931}.
\newblock


\bibitem[Cuomo et~al\mbox{.}(2022)]%
        {cuomo2022scientific}
\bibfield{author}{\bibinfo{person}{Salvatore Cuomo}, \bibinfo{person}{Vincenzo~Schiano Di~Cola}, \bibinfo{person}{Fabio Giampaolo}, \bibinfo{person}{Gianluigi Rozza}, \bibinfo{person}{Maziar Raissi}, {and} \bibinfo{person}{Francesco Piccialli}.} \bibinfo{year}{2022}\natexlab{}.
\newblock \showarticletitle{Scientific machine learning through physics--informed neural networks: Where we are and what’s next}.
\newblock \bibinfo{journal}{\emph{Journal of Scientific Computing}} \bibinfo{volume}{92}, \bibinfo{number}{3} (\bibinfo{year}{2022}), \bibinfo{pages}{88}.
\newblock


\bibitem[Fornberg and Piret(2008)]%
        {fornberg2008stable}
\bibfield{author}{\bibinfo{person}{Bengt Fornberg} {and} \bibinfo{person}{C{\'e}cile Piret}.} \bibinfo{year}{2008}\natexlab{}.
\newblock \showarticletitle{A stable algorithm for flat radial basis functions on a sphere}.
\newblock \bibinfo{journal}{\emph{SIAM Journal on Scientific Computing}} \bibinfo{volume}{30}, \bibinfo{number}{1} (\bibinfo{year}{2008}), \bibinfo{pages}{60--80}.
\newblock


\bibitem[Fornberg and Zuev(2007)]%
        {fornberg2007runge}
\bibfield{author}{\bibinfo{person}{Bengt Fornberg} {and} \bibinfo{person}{Julia Zuev}.} \bibinfo{year}{2007}\natexlab{}.
\newblock \showarticletitle{The Runge phenomenon and spatially variable shape parameters in RBF interpolation}.
\newblock \bibinfo{journal}{\emph{Computers \& Mathematics with Applications}} \bibinfo{volume}{54}, \bibinfo{number}{3} (\bibinfo{year}{2007}), \bibinfo{pages}{379--398}.
\newblock


\bibitem[Gal and Ghahramani(2016)]%
        {gal2016dropout}
\bibfield{author}{\bibinfo{person}{Yarin Gal} {and} \bibinfo{person}{Zoubin Ghahramani}.} \bibinfo{year}{2016}\natexlab{}.
\newblock \showarticletitle{Dropout as a bayesian approximation: Representing model uncertainty in deep learning}. In \bibinfo{booktitle}{\emph{international conference on machine learning}}. PMLR, \bibinfo{pages}{1050--1059}.
\newblock


\bibitem[Geuzaine and Remacle(2009)]%
        {geuzaine2009gmsh}
\bibfield{author}{\bibinfo{person}{Christophe Geuzaine} {and} \bibinfo{person}{Jean-François Remacle}.} \bibinfo{year}{2009}\natexlab{}.
\newblock \showarticletitle{Gmsh: A 3-D finite element mesh generator with built-in pre- and post-processing facilities}.
\newblock \bibinfo{journal}{\emph{Internat. J. Numer. Methods Engrg.}} \bibinfo{volume}{79}, \bibinfo{number}{11} (\bibinfo{year}{2009}), \bibinfo{pages}{1309--1331}.
\newblock
\urldef\tempurl%
\url{https://doi.org/10.1002/nme.2579}
\showDOI{\tempurl}
\showeprint{https://onlinelibrary.wiley.com/doi/pdf/10.1002/nme.2579}


\bibitem[Giles and Pierce(2000)]%
        {giles2000introduction}
\bibfield{author}{\bibinfo{person}{Michael~B Giles} {and} \bibinfo{person}{Niles~A Pierce}.} \bibinfo{year}{2000}\natexlab{}.
\newblock \showarticletitle{An introduction to the adjoint approach to design}.
\newblock \bibinfo{journal}{\emph{Flow, turbulence and combustion}}  \bibinfo{volume}{65} (\bibinfo{year}{2000}), \bibinfo{pages}{393--415}.
\newblock


\bibitem[Griewank and Walther(2008)]%
        {griewank2008evaluating}
\bibfield{author}{\bibinfo{person}{Andreas Griewank} {and} \bibinfo{person}{Andrea Walther}.} \bibinfo{year}{2008}\natexlab{}.
\newblock \bibinfo{booktitle}{\emph{Evaluating derivatives: principles and techniques of algorithmic differentiation}}.
\newblock \bibinfo{publisher}{SIAM}.
\newblock


\bibitem[Hinton et~al\mbox{.}(2012)]%
        {hinton2012imagenet}
\bibfield{author}{\bibinfo{person}{Geoffrey~E Hinton}, \bibinfo{person}{Alex Krizhevsky}, \bibinfo{person}{Ilya Sutskever}, {et~al\mbox{.}}} \bibinfo{year}{2012}\natexlab{}.
\newblock \showarticletitle{Imagenet classification with deep convolutional neural networks}.
\newblock \bibinfo{journal}{\emph{Advances in neural information processing systems}} \bibinfo{volume}{25}, \bibinfo{number}{1106-1114} (\bibinfo{year}{2012}), \bibinfo{pages}{1}.
\newblock


\bibitem[Holl et~al\mbox{.}(2020)]%
        {holl2020learning}
\bibfield{author}{\bibinfo{person}{Philipp Holl}, \bibinfo{person}{Vladlen Koltun}, {and} \bibinfo{person}{Nils Thuerey}.} \bibinfo{year}{2020}\natexlab{}.
\newblock \showarticletitle{Learning to Control PDEs with Differentiable Physics}.
\newblock  (\bibinfo{date}{1} \bibinfo{year}{2020}).
\newblock
\urldef\tempurl%
\url{http://arxiv.org/abs/2001.07457}
\showURL{%
\tempurl}


\bibitem[Hornik et~al\mbox{.}(1989)]%
        {hornik1989multilayer}
\bibfield{author}{\bibinfo{person}{Kurt Hornik}, \bibinfo{person}{Maxwell Stinchcombe}, {and} \bibinfo{person}{Halbert White}.} \bibinfo{year}{1989}\natexlab{}.
\newblock \showarticletitle{Multilayer feedforward networks are universal approximators}.
\newblock \bibinfo{journal}{\emph{Neural networks}} \bibinfo{volume}{2}, \bibinfo{number}{5} (\bibinfo{year}{1989}), \bibinfo{pages}{359--366}.
\newblock


\bibitem[Howell et~al\mbox{.}(2022)]%
        {howell2022dojo}
\bibfield{author}{\bibinfo{person}{Taylor~A Howell}, \bibinfo{person}{Simon Le~Cleac’h}, \bibinfo{person}{J~Zico Kolter}, \bibinfo{person}{Mac Schwager}, {and} \bibinfo{person}{Zachary Manchester}.} \bibinfo{year}{2022}\natexlab{}.
\newblock \showarticletitle{Dojo: A differentiable simulator for robotics}.
\newblock \bibinfo{journal}{\emph{arXiv preprint arXiv:2203.00806}}  \bibinfo{volume}{9} (\bibinfo{year}{2022}).
\newblock


\bibitem[Kansa(1990)]%
        {kansa1990multiquadrics}
\bibfield{author}{\bibinfo{person}{Edward~J Kansa}.} \bibinfo{year}{1990}\natexlab{}.
\newblock \showarticletitle{Multiquadrics—A scattered data approximation scheme with applications to computational fluid-dynamics—I surface approximations and partial derivative estimates}.
\newblock \bibinfo{journal}{\emph{Computers \& Mathematics with applications}} \bibinfo{volume}{19}, \bibinfo{number}{8-9} (\bibinfo{year}{1990}), \bibinfo{pages}{127--145}.
\newblock


\bibitem[Kidger(2022)]%
        {kidger2022neural}
\bibfield{author}{\bibinfo{person}{Patrick Kidger}.} \bibinfo{year}{2022}\natexlab{}.
\newblock \bibinfo{title}{On Neural Differential Equations}.
\newblock
\newblock
\showeprint[arxiv]{2202.02435}~[cs.LG]


\bibitem[Kuhn and Tucker(1951)]%
        {kuhn2013nonlinear}
\bibfield{author}{\bibinfo{person}{Harold~W Kuhn} {and} \bibinfo{person}{Albert~W Tucker}.} \bibinfo{year}{1951}\natexlab{}.
\newblock \showarticletitle{Nonlinear programming}.
\newblock In \bibinfo{booktitle}{\emph{Proceedings of the 2nd Berkeley Symposium on Mathematics, Statistics and Probability}}. \bibinfo{publisher}{University of California Press, Berkeley}, \bibinfo{pages}{481--492}.
\newblock


\bibitem[Lagrange(1853)]%
        {lagrange1853mecanique}
\bibfield{author}{\bibinfo{person}{Joseph~Louis Lagrange}.} \bibinfo{year}{1853}\natexlab{}.
\newblock \bibinfo{booktitle}{\emph{M{\'e}canique analytique}}. Vol.~\bibinfo{volume}{1}.
\newblock \bibinfo{publisher}{Mallet-Bachelier}.
\newblock


\bibitem[Ma et~al\mbox{.}(2021)]%
        {ma2021comparison}
\bibfield{author}{\bibinfo{person}{Yingbo Ma}, \bibinfo{person}{Vaibhav Dixit}, \bibinfo{person}{Michael~J Innes}, \bibinfo{person}{Xingjian Guo}, {and} \bibinfo{person}{Chris Rackauckas}.} \bibinfo{year}{2021}\natexlab{}.
\newblock \showarticletitle{A comparison of automatic differentiation and continuous sensitivity analysis for derivatives of differential equation solutions}. In \bibinfo{booktitle}{\emph{2021 IEEE High Performance Extreme Computing Conference (HPEC)}}. IEEE, \bibinfo{pages}{1--9}.
\newblock


\bibitem[Mowlavi and Nabi(2023)]%
        {mowlavi2023optimal}
\bibfield{author}{\bibinfo{person}{Saviz Mowlavi} {and} \bibinfo{person}{Saleh Nabi}.} \bibinfo{year}{2023}\natexlab{}.
\newblock \showarticletitle{Optimal control of PDEs using physics-informed neural networks}.
\newblock \bibinfo{journal}{\emph{J. Comput. Phys.}}  \bibinfo{volume}{473} (\bibinfo{year}{2023}), \bibinfo{pages}{111731}.
\newblock


\bibitem[Nabi et~al\mbox{.}(2019)]%
        {nabi2019nonlinear}
\bibfield{author}{\bibinfo{person}{Saleh Nabi}, \bibinfo{person}{Piyush Grover}, {and} \bibinfo{person}{CP Caulfield}.} \bibinfo{year}{2019}\natexlab{}.
\newblock \showarticletitle{Nonlinear optimal control strategies for buoyancy-driven flows in the built environment}.
\newblock \bibinfo{journal}{\emph{Computers \& Fluids}}  \bibinfo{volume}{194} (\bibinfo{year}{2019}), \bibinfo{pages}{104313}.
\newblock


\bibitem[Nimier-David et~al\mbox{.}(2019)]%
        {nimier2019mitsuba}
\bibfield{author}{\bibinfo{person}{Merlin Nimier-David}, \bibinfo{person}{Delio Vicini}, \bibinfo{person}{Tizian Zeltner}, {and} \bibinfo{person}{Wenzel Jakob}.} \bibinfo{year}{2019}\natexlab{}.
\newblock \showarticletitle{Mitsuba 2: A retargetable forward and inverse renderer}.
\newblock \bibinfo{journal}{\emph{ACM Transactions on Graphics (TOG)}} \bibinfo{volume}{38}, \bibinfo{number}{6} (\bibinfo{year}{2019}), \bibinfo{pages}{1--17}.
\newblock


\bibitem[Nzoyem({[n.\,d.]})]%
        {nzoyem2023updec}
\bibfield{author}{\bibinfo{person}{Roussel Nzoyem}.} \bibinfo{year}{[n.\,d.]}\natexlab{}.
\newblock \bibinfo{title}{Updec}.
\newblock
\newblock
\urldef\tempurl%
\url{https://github.com/ddrous/Updec/}
\showURL{%
\tempurl}
\newblock
\shownote{Third Updec pre-release version.}.


\bibitem[Orr et~al\mbox{.}(1996)]%
        {orr1996introduction}
\bibfield{author}{\bibinfo{person}{Mark~JL Orr} {et~al\mbox{.}}} \bibinfo{year}{1996}\natexlab{}.
\newblock \bibinfo{title}{Introduction to radial basis function networks}.
\newblock
\newblock


\bibitem[Park and Sandberg(1991)]%
        {park1991universal}
\bibfield{author}{\bibinfo{person}{Jooyoung Park} {and} \bibinfo{person}{Irwin~W Sandberg}.} \bibinfo{year}{1991}\natexlab{}.
\newblock \showarticletitle{Universal approximation using radial-basis-function networks}.
\newblock \bibinfo{journal}{\emph{Neural computation}} \bibinfo{volume}{3}, \bibinfo{number}{2} (\bibinfo{year}{1991}), \bibinfo{pages}{246--257}.
\newblock


\bibitem[Pontryagin et~al\mbox{.}(1962)]%
        {pontryagin1962maximum}
\bibfield{author}{\bibinfo{person}{LS Pontryagin}, \bibinfo{person}{VG Boltyanskii}, \bibinfo{person}{RV Gamkrelidze}, {and} \bibinfo{person}{EF Mishchenko}.} \bibinfo{year}{1962}\natexlab{}.
\newblock \showarticletitle{The maximum principle}.
\newblock \bibinfo{journal}{\emph{The Mathematical Theory of Optimal Processes. New York: John Wiley and Sons}} (\bibinfo{year}{1962}).
\newblock


\bibitem[Powell(1981)]%
        {powell1981approximation}
\bibfield{author}{\bibinfo{person}{Michael James~David Powell}.} \bibinfo{year}{1981}\natexlab{}.
\newblock \bibinfo{booktitle}{\emph{Approximation theory and methods}}.
\newblock \bibinfo{publisher}{Cambridge university press}.
\newblock


\bibitem[Raissi et~al\mbox{.}(2019)]%
        {raissi2019physics}
\bibfield{author}{\bibinfo{person}{Maziar Raissi}, \bibinfo{person}{Paris Perdikaris}, {and} \bibinfo{person}{George~E Karniadakis}.} \bibinfo{year}{2019}\natexlab{}.
\newblock \showarticletitle{Physics-informed neural networks: A deep learning framework for solving forward and inverse problems involving nonlinear partial differential equations}.
\newblock \bibinfo{journal}{\emph{Journal of Computational physics}}  \bibinfo{volume}{378} (\bibinfo{year}{2019}), \bibinfo{pages}{686--707}.
\newblock


\bibitem[Raissi et~al\mbox{.}(2018)]%
        {raissi2018hidden}
\bibfield{author}{\bibinfo{person}{Maziar Raissi}, \bibinfo{person}{Alireza Yazdani}, {and} \bibinfo{person}{George~Em Karniadakis}.} \bibinfo{year}{2018}\natexlab{}.
\newblock \showarticletitle{Hidden fluid mechanics: A Navier-Stokes informed deep learning framework for assimilating flow visualization data}.
\newblock \bibinfo{journal}{\emph{arXiv preprint arXiv:1808.04327}} (\bibinfo{year}{2018}).
\newblock


\bibitem[Rumelhart et~al\mbox{.}(1986)]%
        {rumelhart1986learning}
\bibfield{author}{\bibinfo{person}{David~E Rumelhart}, \bibinfo{person}{Geoffrey~E Hinton}, {and} \bibinfo{person}{Ronald~J Williams}.} \bibinfo{year}{1986}\natexlab{}.
\newblock \showarticletitle{Learning representations by back-propagating errors}.
\newblock \bibinfo{journal}{\emph{nature}} \bibinfo{volume}{323}, \bibinfo{number}{6088} (\bibinfo{year}{1986}), \bibinfo{pages}{533--536}.
\newblock


\bibitem[Sabne(2020)]%
        {sabne2020xla}
\bibfield{author}{\bibinfo{person}{Amit Sabne}.} \bibinfo{year}{2020}\natexlab{}.
\newblock \showarticletitle{Xla: Compiling machine learning for peak performance}.
\newblock  (\bibinfo{year}{2020}).
\newblock


\bibitem[Schoenholz and Cubuk(2019)]%
        {schoenholz2019jax}
\bibfield{author}{\bibinfo{person}{Samuel~S Schoenholz} {and} \bibinfo{person}{Ekin~D Cubuk}.} \bibinfo{year}{2019}\natexlab{}.
\newblock \showarticletitle{Jax md: End-to-end differentiable, hardware accelerated, molecular dynamics in pure python}.
\newblock  (\bibinfo{year}{2019}).
\newblock


\bibitem[Shahane et~al\mbox{.}(2020)]%
        {shahane2020ahighorder}
\bibfield{author}{\bibinfo{person}{Shantanu Shahane}, \bibinfo{person}{Anand Radhakrishnan}, {and} \bibinfo{person}{Surya~Pratap Vanka}.} \bibinfo{year}{2020}\natexlab{}.
\newblock \showarticletitle{A High-Order Accurate Meshless Method for Solution of Incompressible Fluid Flow Problems}.
\newblock  (\bibinfo{date}{10} \bibinfo{year}{2020}).
\newblock
\urldef\tempurl%
\url{https://doi.org/10.1016/j.jcp.2021.110623}
\showDOI{\tempurl}


\bibitem[Shahane and Vanka(2023)]%
        {shahane2023semi}
\bibfield{author}{\bibinfo{person}{Shantanu Shahane} {and} \bibinfo{person}{Surya~Pratap Vanka}.} \bibinfo{year}{2023}\natexlab{}.
\newblock \showarticletitle{A semi-implicit meshless method for incompressible flows in complex geometries}.
\newblock \bibinfo{journal}{\emph{J. Comput. Phys.}}  \bibinfo{volume}{472} (\bibinfo{year}{2023}), \bibinfo{pages}{111715}.
\newblock


\bibitem[Shen et~al\mbox{.}(2023)]%
        {shen2023differentiable}
\bibfield{author}{\bibinfo{person}{Chaopeng Shen}, \bibinfo{person}{Alison~P Appling}, \bibinfo{person}{Pierre Gentine}, \bibinfo{person}{Toshiyuki Bandai}, \bibinfo{person}{Hoshin Gupta}, \bibinfo{person}{Alexandre Tartakovsky}, \bibinfo{person}{Marco Baity-Jesi}, \bibinfo{person}{Fabrizio Fenicia}, \bibinfo{person}{Daniel Kifer}, \bibinfo{person}{Li Li}, {et~al\mbox{.}}} \bibinfo{year}{2023}\natexlab{}.
\newblock \showarticletitle{Differentiable modelling to unify machine learning and physical models for geosciences}.
\newblock \bibinfo{journal}{\emph{Nature Reviews Earth \& Environment}} (\bibinfo{year}{2023}), \bibinfo{pages}{1--16}.
\newblock


\bibitem[Tolstykh(2000)]%
        {tolstykh2000using}
\bibfield{author}{\bibinfo{person}{Andrei~I Tolstykh}.} \bibinfo{year}{2000}\natexlab{}.
\newblock \showarticletitle{On using RBF-based differencing formulas for unstructured and mixed structured-unstructured grid calculations}. In \bibinfo{booktitle}{\emph{Proceedings of the 16th IMACS world congress}}, Vol.~\bibinfo{volume}{228}. Lausanne, \bibinfo{pages}{4606--4624}.
\newblock


\bibitem[Tr{\'e}lat(2005)]%
        {trelat2005controle}
\bibfield{author}{\bibinfo{person}{Emmanuel Tr{\'e}lat}.} \bibinfo{year}{2005}\natexlab{}.
\newblock \bibinfo{booktitle}{\emph{Contr{\^o}le optimal: th{\'e}orie \& applications}}. Vol.~\bibinfo{volume}{36}.
\newblock \bibinfo{publisher}{Vuibert Paris}.
\newblock


\bibitem[Tsay et~al\mbox{.}(2020)]%
        {tsay2020modeling}
\bibfield{author}{\bibinfo{person}{Calvin Tsay}, \bibinfo{person}{Fernando Lejarza}, \bibinfo{person}{Mark~A Stadtherr}, {and} \bibinfo{person}{Michael Baldea}.} \bibinfo{year}{2020}\natexlab{}.
\newblock \showarticletitle{Modeling, state estimation, and optimal control for the US COVID-19 outbreak}.
\newblock \bibinfo{journal}{\emph{Scientific reports}} \bibinfo{volume}{10}, \bibinfo{number}{1} (\bibinfo{year}{2020}), \bibinfo{pages}{10711}.
\newblock


\bibitem[Van~Rossum et~al\mbox{.}(2007)]%
        {van2007python}
\bibfield{author}{\bibinfo{person}{Guido Van~Rossum} {et~al\mbox{.}}} \bibinfo{year}{2007}\natexlab{}.
\newblock \showarticletitle{Python Programming Language.}. In \bibinfo{booktitle}{\emph{USENIX annual technical conference}}, Vol.~\bibinfo{volume}{41}. Santa Clara, CA, \bibinfo{pages}{1--36}.
\newblock


\bibitem[Vaswani et~al\mbox{.}(2017)]%
        {vaswani2017attention}
\bibfield{author}{\bibinfo{person}{Ashish Vaswani}, \bibinfo{person}{Noam Shazeer}, \bibinfo{person}{Niki Parmar}, \bibinfo{person}{Jakob Uszkoreit}, \bibinfo{person}{Llion Jones}, \bibinfo{person}{Aidan~N Gomez}, \bibinfo{person}{{\L}ukasz Kaiser}, {and} \bibinfo{person}{Illia Polosukhin}.} \bibinfo{year}{2017}\natexlab{}.
\newblock \showarticletitle{Attention is all you need}.
\newblock \bibinfo{journal}{\emph{Advances in neural information processing systems}}  \bibinfo{volume}{30} (\bibinfo{year}{2017}).
\newblock


\bibitem[Werbos(1974)]%
        {werbos1974beyond}
\bibfield{author}{\bibinfo{person}{Paul Werbos}.} \bibinfo{year}{1974}\natexlab{}.
\newblock \showarticletitle{Beyond regression: New tools for prediction and analysis in the behavioral sciences}.
\newblock \bibinfo{journal}{\emph{PhD thesis, Committee on Applied Mathematics, Harvard University, Cambridge, MA}} (\bibinfo{year}{1974}).
\newblock


\bibitem[Werbos(1990)]%
        {werbos1990backpropagation}
\bibfield{author}{\bibinfo{person}{Paul~J Werbos}.} \bibinfo{year}{1990}\natexlab{}.
\newblock \showarticletitle{Backpropagation through time: what it does and how to do it}.
\newblock \bibinfo{journal}{\emph{Proc. IEEE}} \bibinfo{volume}{78}, \bibinfo{number}{10} (\bibinfo{year}{1990}), \bibinfo{pages}{1550--1560}.
\newblock


\bibitem[Zamolo and Nobile(2019)]%
        {zamolo2019solution}
\bibfield{author}{\bibinfo{person}{R. Zamolo} {and} \bibinfo{person}{E. Nobile}.} \bibinfo{year}{2019}\natexlab{}.
\newblock \showarticletitle{Solution of incompressible fluid flow problems with heat transfer by means of an efficient RBF-FD meshless approach}.
\newblock \bibinfo{journal}{\emph{Numerical Heat Transfer, Part B: Fundamentals}}  \bibinfo{volume}{75} (\bibinfo{date}{1} \bibinfo{year}{2019}), \bibinfo{pages}{19--42}.
\newblock
Issue 1.
\showISSN{15210626}
\urldef\tempurl%
\url{https://doi.org/10.1080/10407790.2019.1580048}
\showDOI{\tempurl}


\end{thebibliography}

%% Appendix
% \appendix

\end{document}